\documentclass[twocolumn]{svjour3}          
\smartqed  

\usepackage{graphicx}
\usepackage{balance}  
\usepackage[linesnumbered,ruled,noend]{algorithm2e}
\usepackage{graphicx, caption, subcaption}
\usepackage{multirow}
\usepackage{multicol}
\usepackage{booktabs}
\usepackage{color}
\usepackage{hyperref}
\usepackage{amsmath}
\usepackage{cancel}
\usepackage{ifthen}

\usepackage{amsthm}
\usepackage{makecell}
\usepackage{comment}
\usepackage{cite}
\usepackage{url}

\usepackage{breakurl}

\newcommand{\squishlist}{ 
   \begin{list}{$\bullet$}
    { \setlength{\itemsep}{0pt}      \setlength{\parsep}{3pt} 
      \setlength{\topsep}{3pt}       \setlength{\partopsep}{0pt}
      \setlength{\leftmargin}{1.5em} \setlength{\labelwidth}{1em}
      \setlength{\labelsep}{0.5em} } }

\newcommand{\squishend}{
    \end{list}  }

\newboolean{techreport}
\setboolean{techreport}{true}
\newboolean{color}
\setboolean{color}{false}

\usepackage[normalem]{ulem}
\usepackage{xcolor}

\newcommand\bluesout{\bgroup\markoverwith{\textcolor{blue}{\rule[0.5ex]{2pt}{0.4pt}}}\ULon}

\ifthenelse{\boolean{color}}{

}{

}

\newcommand{\chgd}[1]{{\color{black}#1}} 
\newcommand{\revision}[1]{{\color{black}#1}}

\begin{document}

\title{Data Collection and Quality Challenges in Deep Learning: A Data-Centric AI Perspective\thanks{This article extends tutorials
the authors delivered at the VLDB 2020\,\cite{DBLP:journals/pvldb/Whang020} and KDD 2021\,\cite{DBLP:conf/kdd/0001RSW21} conferences.}}

\author{Steven Euijong Whang        \and
        Yuji Roh \and
        Hwanjun Song \and
        Jae-Gil Lee 
}


\institute{S. E. Whang \at
              KAIST\\
              \email{swhang@kaist.ac.kr}           
            \and
            Y. Roh \at
              KAIST\\
              \email{yuji.roh@kaist.ac.kr} 
           \and
            H. Song \at
              Naver AI Lab\\
              \email{hwanjun.song@navercorp.com} 
           \and
           J. Lee \at
              KAIST\\
              \email{jaegil@kaist.ac.kr} 
}

\date{Received: date / Accepted: date}

\sloppy

\maketitle

\begin{abstract}
\chgd{Data-centric AI is at the center of a fundamental shift in software engineering where machine learning becomes the new software, powered by big data and computing infrastructure. Here software engineering needs to be re-thought where data becomes a first-class citizen on par with code. One striking observation is that a significant portion of the machine learning process is spent on data preparation.} Without good data, even the best machine learning algorithms cannot perform well. As a result, data-centric AI practices are now becoming mainstream. Unfortunately, many datasets in the real world are small, dirty, biased, and even poisoned. In this survey, we study the research landscape for data collection and data quality primarily for deep learning applications. Data collection is important because there is lesser need for feature engineering for recent deep learning approaches, but instead more need for large amounts of data. \chgd{For data quality, we study data validation, cleaning, and integration techniques. Even if the data cannot be fully cleaned, we can still cope with imperfect data during model training using robust model training techniques.} 
In addition, while bias and fairness have been less studied in traditional data management research, these issues become essential topics in modern machine learning applications. 
We thus study fairness measures and unfairness mitigation techniques that can be applied before, during, or after model training. We believe that the data management community is well poised to solve these problems.
\end{abstract}

\section{Overview}

Deep learning is widely used to glean knowledge from massive amounts of data. There is a wide range of applications including natural language understanding, healthcare, self-driving cars, and more. Deep learning has become so prevalent thanks to its excellent performance with the availability of big data and powerful computing infrastructure. According to the IDC\,\cite{idc}, the amount of data worldwide is projected to grow exponentially to 175 zettabytes (ZB) by 2025. In addition, powerful GPUs and TPUs enable software to have superhuman performances in various tasks.

\begin{figure*}[t]
  \centering
  \includegraphics[width=0.75\textwidth]{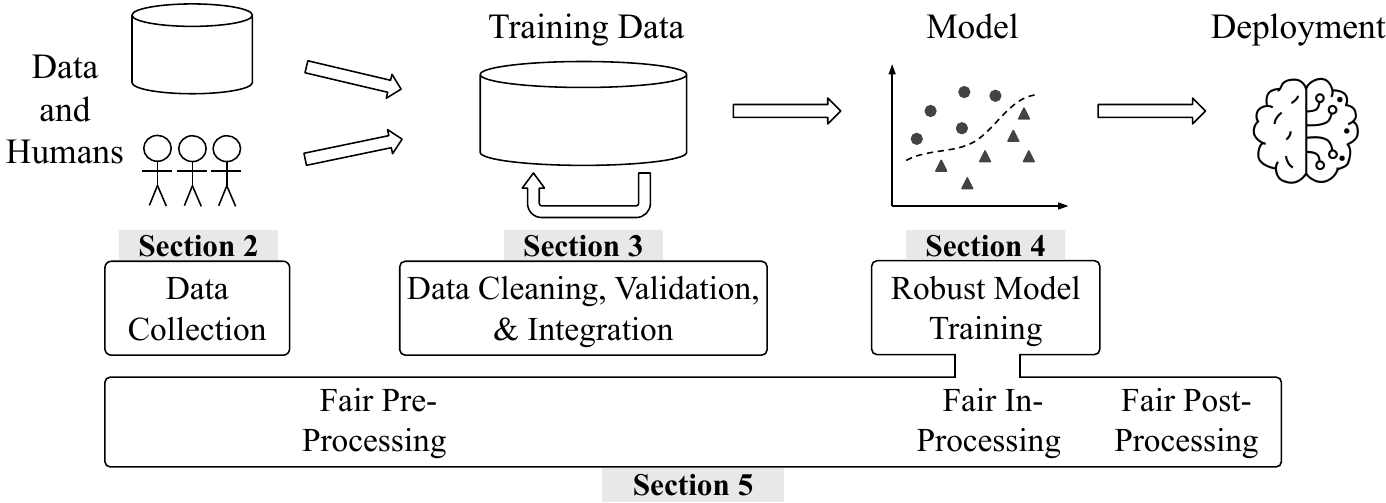}
  \vspace{-0.2cm}
  \caption{Deep learning challenges from a data-centric AI perspective. Data collection and quality issues cannot be resolved in a single step, but throughout the entire machine learning process. This survey thus focuses on the breadth of available techniques.}
  \label{fig:e2e}
  \vspace{-0.3cm}
\end{figure*}

\chgd{We are going through a fundamental paradigm shift in software engineering where machine learning becomes the new software (referred to as Software 2.0\,\cite{karpathi}).} Conventional software engineering involves designing, implementing, and debugging code. In comparison, machine learning starts with data and trains a function on the data. It is known that data preparation is an expensive step in end-to-end machine learning. \chgd{In particular, collecting data, cleaning it, and making it suitable for machine learning training takes 45\%\,\cite{datanami} or even 80--90\%\,\cite{DBLP:journals/debu/Stonebraker19,crowdflower} of the entire time.} In addition, the code on a machine learning platform (e.g., PyTorch\,\cite{NEURIPS2019_9015}) is high level and thus requires significantly fewer lines compared to conventional software. Finally, the trained model may need to be continuously improved with hyperparameter tuning. 
This entire process from data preparation to model deployment is widely viewed as a new software engineering paradigm, and companies have been actively developing open source \cite{DBLP:conf/kdd/BaylorBCFFHHIJK17,DBLP:conf/sigmod/ChenCDD0HKMMNOP20} and proprietary Software 2.0 systems\,\cite{DBLP:conf/cidr/BoehmADGIKLPR20,DBLP:conf/sigmod/AgrawalABBGGLMP19,DBLP:conf/icse/AmershiBBDGKNN019,DBLP:conf/hpca/HazelwoodBBCDDF18,michelangelo,Schelter2017AutomaticallyTM}.
Solving data issues is increasingly becoming critical in machine learning research.

While data collection and quality issues are important, machine learning research has mainly focused on training algorithms instead of the data. \chgd{According to\,\cite{DBLP:journals/debu/Stonebraker19}, a common complaint in the industry is that research institutions spend 90\% of their machine learning efforts on algorithms and 10\% on data preparation, although based on the amounts of time spent, the numbers should be 10\% and 90\% the other way.}

\chgd{At the same time, many companies are promising to use responsible and data-centric AI practices.} For example, Google\,\cite{google} says that AI has a significant potential to help solve challenging problems, but it is important to develop responsibly. Microsoft\,\cite{microsoft} pledges to advance AI using ethical principles that put people first. Other companies make similar statements\,\cite{ibm,samsung}. More recently, data-centric AI\,\cite{datacentricai} is becoming critical where the primary goal is not to improve the model training algorithm, but to improve the data pre-processing for better model accuracy.

These trends motivate us to investigate data collection and quality challenges for deep learning from a data-centric AI perspective. Figure~\ref{fig:e2e} shows a simplified end-to-end process starting from data collection to model deployment. Deep learning systems are more complicated in practice\,\cite{DBLP:conf/nips/SculleyHGDPECYC15,DBLP:journals/sigmod/PolyzotisRWZ18}, and we only show the essential steps. 
The first topic we cover is \emph{data collection}. In comparison to traditional machine learning, in deep learning feature engineering is less of a concern, but there is instead a need for large amounts of training data. Unfortunately, many industries do not adopt deep learning simply because of the lack of data and the lack of explainability of the trained models.
The second topic is \emph{data cleaning and validation}. While there is a vast literature on data cleaning, unfortunately not all the techniques directly benefit deep learning accuracy\,\cite{cleanmlli}. In addition, there are recent deep learning issues including data poisoning that needs to be addressed, especially by the data management community. \chgd{Data poisoning is becoming a significant threat as attackers generate data with a malicious intent to reduce the model accuracy of AI applications. In response, there is a branch of research called data sanitization where the goal is to defend against such attacks.}
The third topic is \emph{robust model training}. Even after we carefully validate and clean our data, the data quality may still be problematic because there is no guarantee that we fixed all the data problems. Hence, we may still need to cope with dirty, missing, or even poisoned data in model training. Fortunately, there are various robust training techniques\,\cite{DBLP:conf/icml/SongK019} available.
The fourth topic is \emph{fair model training}. 
\chgd{Traditional research on data management has not focused on bias and fairness issues. However, in addition to cleaning and validating data to improve model accuracy, also showing fairness against biased data is becoming essential for responsible AI. In fact, many data validation works now mention that supporting AI ethics including fairness is an important future research direction\,\cite{DBLP:journals/debu/BiessmannGRL021}. Model fairness research\,\cite{barocas-hardt-narayanan,DBLP:conf/pods/Venkatasubramanian19,DBLP:journals/cacm/ChouldechovaR20,DBLP:journals/corr/abs-1908-09635} largely consists of fairness measures and unfairness mitigation before, during, or after model training.} \revision{Recent studies are now addressing model fairness and robustness together due to their close relationship where data bias and noise may affect each other in the same training data\,\cite{DBLP:journals/corr/abs-2002-10234,roh2021sample,DBLP:conf/pkdd/SolansB020, DBLP:conf/fat/KhaniL21}.} 


\begin{figure}[ht]
\vspace{-0.2cm}
\center
  \includegraphics[width=\columnwidth]{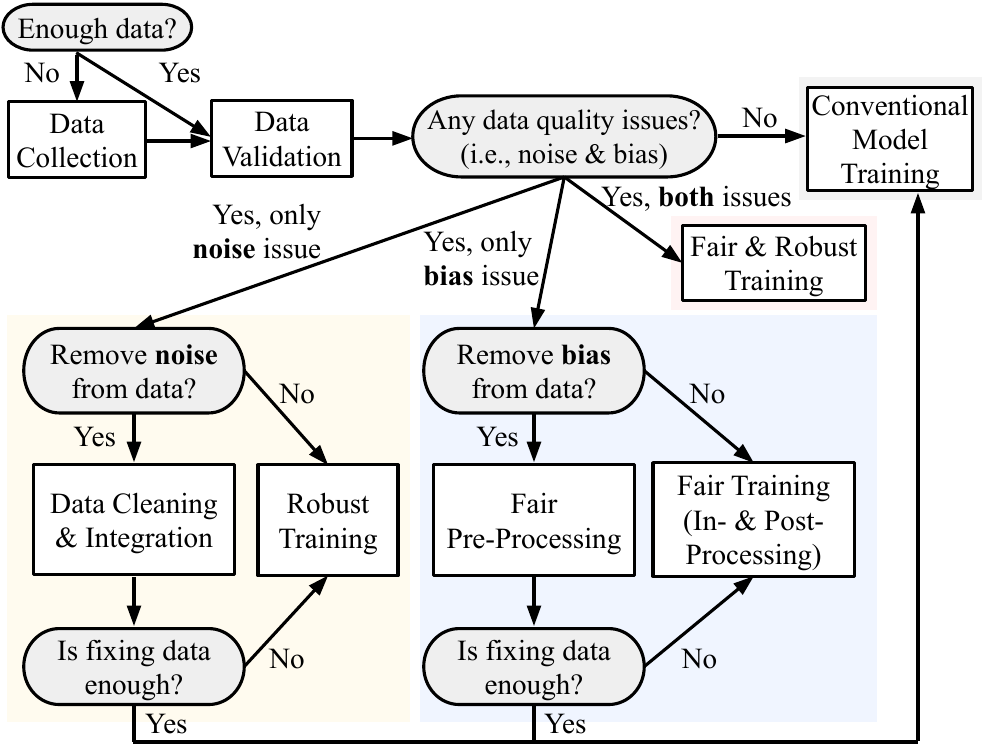}
  \vspace{-0.2cm}
  \caption{\revision{Decision tree on how data-centric AI techniques connect with each other in one workflow.}}
  \vspace{-0.3cm}
  \label{fig:decisiontree}
\end{figure}

\chgd{While the coverage of this survey is broad, we believe it is important to have a birds-eye view of data issues in the entire deep learning process in order to advance data-centric AI.} Each subtopic is not only substantial, but studied by different communities. Data collection, cleaning, and validation have been traditionally studied in the data management community. Robust model training is a central topic in the machine learning and security communities while fair model training is a popular topic in the machine learning and fairness communities. Both fairness and robustness topics are increasingly being researched in the data management community as well because they are closely related to the input data. \chgd{Data-centric AI is a nascent field that cannot be covered by solving just one of these areas either, but instead will ultimately need an orchestration within a holistic framework.} Our contribution is thus to connect these related topics together at a high level with a focus on recent and significant works. Table~\ref{tbl:alltechniques} shows a taxonomy of the techniques covered in this survey. \chgd{Figure~\ref{fig:decisiontree} shows a decision tree of how the techniques connect with each other in one workflow.} Our work targets researchers and practitioners who need a starting point of understanding how data plays a key role in data-centric AI.

\begin{table*}[t]
  \centering
  \scalebox{0.9}{
  \begin{tabular}{@{\hspace{6pt}}cc@{\hspace{6pt}}c@{\hspace{6pt}}ccc}
    \toprule
     {\bf Operation} & {\bf Category} & {\bf Section} & {\bf Data Type} & {\bf Technique} & {\bf Key References} \\
     \midrule
      \multirow{7}[8]{*}{Data Collection} & \multirow{3}[4]{*}{Data Acquisition} & \multirow{3}[4]{*}{\ref{sec:dataacquisition}} & \multirow{3}[4]{*}{All} & Data discovery & \,\cite{DBLP:conf/cidr/TerrizzanoSRC15,DBLP:conf/icde/FernandezAKYMS18,DBLP:journals/pvldb/CafarellaHLMYWW18,DBLP:conf/sigmod/HalevyKNOPRW16,DBLP:conf/www/BrickleyBN19,DBLP:conf/sigmod/ZhangI20,DBLP:journals/debu/MillerNZCPA18}\\
      \cmidrule(lr){5-6}
      & & & & Data augmentation & \,\cite{DBLP:conf/nips/GoodfellowPMXWOCB14,DBLP:journals/corr/Goodfellow17,DBLP:journals/corr/abs-1807-04720,DBLP:conf/nips/RatnerEHDR17,DBLP:journals/corr/abs-2004-03264,DBLP:conf/cvpr/CubukZMVL19,DBLP:conf/iclr/ZhangCDL18,modelpatching} \\ 
      \cmidrule(lr){5-6}
      & & & & Data generation & \,\cite{amazonmechanicalturk,DBLP:conf/pkdd/PelekisNTST13,DBLP:conf/aaai/KimLHS19,DBLP:conf/iros/TobinFRSZA17,DBLP:conf/cvpr/TremblayPABJATC18} \\
      \cmidrule(lr){2-6}
      & \multirow{3}[4]{*}{Data Labeling} & \multirow{3}[4]{*}{\ref{sec:datalabeling}} & \multirow{3}[4]{*}{All} &  Utilize existing labels & \,\cite{DBLP:journals/kais/TrigueroGH15,zhu2008,Yarowsky:1995:UWS:981658.981684}\\
      \cmidrule(lr){5-6}
      & & & & Manual labeling & \,\cite{amazonsagemaker,gcp,DBLP:series/synthesis/2012Settles,DBLP:reference/sp/2015rsh}\\
      \cmidrule(lr){5-6}
      & & & & Automatic labeling & \,\cite{DBLP:journals/vldb/RatnerBEFWR20,Ratner:2017:SRT:3173074.3173077,DBLP:conf/sigmod/BachRLLSXSRHAKR19,DBLP:journals/pvldb/VarmaR18}\\
      \cmidrule(lr){2-6}
      & Improve Existing Data & \ref{sec:improvingexisting} & {Tabular} &  Improve labels and model & \,\cite{DBLP:conf/kdd/ShengPI08,DBLP:journals/tkde/PanY10,tfhub} \\
      \midrule
      \multirow{9}[10]{*}{\makecell{Data Validation,\\ Cleaning, and\\Integration}} & \multirow{4}[5]{*}{Data Validation}  & \multirow{4}[5]{*}{\ref{sec:datavalidation}} & \multirow{4}[5]{*}{Tabular \& Image} & Visualization & \,\cite{DBLP:journals/debu/LeeP18,DBLP:journals/pvldb/VartakRMPP15} \\
      \cmidrule(lr){5-6}
      & & & & False discovery control & \,\cite{DBLP:conf/sigmod/ZhaoSZBUK17,https://doi.org/10.1111/j.1467-9868.2007.00643.x} \\
      \cmidrule(lr){5-6}
      & & & & Schema-based validation & \,\cite{breck2019data,DBLP:conf/kdd/BaylorBCFFHHIJK17,Polyzotis:2017:DMC:3035918.3054782} \\
      \cmidrule(lr){5-6}
      & & & & New functionalities & \,\cite{DBLP:conf/cidr/GrafbergerSS21,DBLP:journals/pvldb/SchelterLSCBG18,DBLP:conf/icde/SchelterGSRKTBL19,DBLP:conf/edbt/RedyukKMS21,DBLP:conf/edbt/SchelterRB21,DBLP:conf/cidr/MelgarDGGHJKL0L21,DBLP:conf/cidr/GrafbergerSS21,DBLP:journals/debu/XinPTWGHJP21}\\
      \cmidrule(lr){2-6}
      & \multirow{2}[2]{*}{Data Cleaning} & \multirow{2}[2]{*}{\ref{sec:datacleaning}} & \multirow{2}[2]{*}{Tabular \& Image} & Data-only cleaning & \,\cite{DBLP:books/acm/IlyasC19,DBLP:journals/pvldb/RekatsinasCIR17}\\
      \cmidrule(lr){5-6}
      & & & & ML-aware cleaning & \,\cite{cleanmlli,DBLP:conf/kdd/DongR19,DBLP:journals/pvldb/KrishnanWWFG16,DBLP:journals/pvldb/DolatshahTWP18,DBLP:journals/debu/RenggliRGK0021,DBLP:journals/pvldb/KarlasLWGC0020,DBLP:journals/debu/NeutatzCA021}\\
      \cmidrule(lr){2-6}
      & \multirow{2}[2]{*}{Data Sanitization} & \multirow{2}[2]{*}{\ref{sec:datasanitization}} & \multirow{2}[2]{*}{Tabular \& Image} & Data poisoning & \,\cite{DBLP:conf/nips/ShafahiHNSSDG18,DBLP:conf/icml/ZhuHLTSG19} \\
      \cmidrule(lr){5-6}
      & & & & Poisoning defenses & \,\cite{DBLP:conf/sp/CretuSLSK08,DBLP:journals/corr/abs-1811-00741,DBLP:journals/air/HodgeA04,DBLP:journals/corr/abs-1802-03041} \\
      \cmidrule(lr){2-6}
      & Data Integration & \ref{sec:dataintegration} & {Multimodal} & Multimodal integration & \,\cite{DBLP:journals/pami/BaltrusaitisAM19,DBLP:journals/debu/StonebrakerI18,DBLP:books/daglib/0029346} \\
      \midrule
      \multirow{4}[5]{*}{\makecell{Robust Model\\Training}} & Noisy Features & \ref{sec:noisy_feature} & {Image} & Adversarial learning & \,\cite{DBLP:journals/corr/ReedLASER14,DBLP:conf/cvpr/PatriniRMNQ17,DBLP:conf/nips/ChangLM17} \\
      \cmidrule(lr){2-6}
      & Missing Features & \ref{sec:missing_feature} & {All} & Data imputation & \,\cite{CheEtAl_nature_sr18} \\
      \cmidrule(lr){2-6}
      & Noisy Labels & \ref{sec:noisy_label} & {Image} & Robust learning & \,\cite{DBLP:conf/icml/SongK019,DBLP:conf/iclr/LiSH20,DBLP:conf/nips/MalachS17,DBLP:conf/icml/JiangZLLF18,DBLP:conf/nips/HanYYNXHTS18} \\
      \cmidrule(lr){2-6}
      & Missing Labels & \ref{sec:missing_label} & {Image} & Semi-supervised learning & \,\cite{DBLP:conf/nips/TarvainenV17,DBLP:conf/nips/BerthelotCGPOR19} \\ 
      \midrule
      \multirow{5}[6]{*}{\makecell{Measuring\\Fairness}} & \multirow{3}[4]{*}{Statistical Fairness} & \multirow{3}[4]{*}{\ref{sec:measuringfairness}} & \multirow{3}[4]{*}{All} & Independence criteria ($\hat{Y} \bot Z$) & \,\cite{DBLP:conf/kdd/FeldmanFMSV15,DBLP:conf/innovations/DworkHPRZ12} \\
      \cmidrule(lr){5-6}
      & & & & Separation criteria ($\hat{Y} \bot Z | Y$) & \,\cite{DBLP:conf/nips/HardtPNS16,DBLP:conf/www/ZafarVGG17,berk2017fairness} \\
      \cmidrule(lr){5-6}
      & & & & Sufficiency criteria ($Y \bot Z | \hat{Y}$) & \,\cite{DBLP:journals/bigdata/Chouldechova17,2016COMPASRS,berk2017fairness} \\
      \cmidrule(lr){2-6}
      & \multirow{2}[2]{*}{Other Fairness} & \multirow{2}[2]{*}{\ref{sec:measuringfairness}} & \multirow{2}[2]{*}{All} & Individual fairness & \,\cite{DBLP:conf/innovations/DworkHPRZ12}\\
      \cmidrule(lr){5-6}
      & & & & Causal fairness & \,\cite{10.5555/3294771.3294834,NIPS2017_6995,Zhang2018FairnessID,Nabi2018FairIO,10.1145/3308558.3313559}\\
      \midrule
      \multirow{10}[11]{*}{\makecell{Unfairness\\Mitigation}} & \multirow{3}[4]{*}{Pre-processing} & \multirow{3}[4]{*}{\ref{sec:unfairnessmitigation}} & \multirow{3}[4]{*}{Tabular \& Image} & Repair data & \,\cite{DBLP:conf/kdd/FeldmanFMSV15,DBLP:conf/sigmod/SalimiRHS19,DBLP:journals/kais/KamiranC11} \\
      \cmidrule(lr){5-6}
      & & & & Generate data & \,\cite{DBLP:conf/icml/ChoiGSSE20,DBLP:conf/bigdataconf/XuYZW18} \\
      \cmidrule(lr){5-6}
      & & & & Acquire data & \,\cite{DBLP:conf/sigmod/TaeW21,DBLP:conf/nips/ChenJS18,DBLP:conf/icde/AsudehJJ19} \\
      \cmidrule(lr){2-6}
      & \multirow{3}[4]{*}{In-processing} & \multirow{3}[4]{*}{\ref{sec:unfairnessmitigation}} & \multirow{3}[4]{*}{Tabular} & Fairness constraints & \,\cite{DBLP:conf/aistats/ZafarVGG17,DBLP:conf/icml/AgarwalBD0W18,DBLP:conf/pkdd/KamishimaAAS12} \\
      \cmidrule(lr){5-6}
      & & & & Adversarial training & \,\cite{DBLP:conf/aies/ZhangLM18,DBLP:conf/alt/CotterJS19,DBLP:journals/corr/abs-2002-10234} \\
      \cmidrule(lr){5-6}
      & & & & Adaptive reweighting & \,\cite{DBLP:conf/iclr/Roh0WS21,DBLP:conf/sigmod/ZhangCAN21,DBLP:conf/aistats/JiangN20,DBLP:conf/cikm/IosifidisN19} \\
      \cmidrule(lr){2-6}
      & Post-processing & \ref{sec:unfairnessmitigation} & {Tabular} & Fix predictions & \,\cite{DBLP:conf/nips/HardtPNS16,DBLP:conf/nips/ChzhenDHOP19,DBLP:conf/nips/PleissRWKW17}\\
      \cmidrule(lr){2-6}
      & \multirow{3}[4]{*}{\makecell{Convergence with\\ Robustness}} & \multirow{3}[4]{*}{\ref{sec:robustandfair}} & \multirow{3}[4]{*}{Tabular \& Image} & Fairness-oriented & \,\cite{DBLP:conf/nips/WangGNCGJ20,DBLP:conf/nips/LamyZ19,DBLP:conf/icml/HashimotoSNL18,DBLP:conf/nips/LahotiBCLPT0C20} \\
      \cmidrule(lr){5-6}
      & & & & Robustness-oriented & \,\cite{DBLP:conf/fat/ZhangD21,DBLP:conf/fat/KhaniL21,DBLP:conf/icml/XuLLJT21} \\
      \cmidrule(lr){5-6}
      & & & & Equal Mergers & \,\cite{DBLP:journals/corr/abs-2002-10234,roh2021sample,DBLP:conf/fat/WangLL21,DBLP:conf/pkdd/SolansB020} \\
    \bottomrule
  \end{tabular}
  }
  \vspace{-0.1cm}
  \caption{\chgd{Taxonomy of data collection and quality techniques for deep learning.}}
  \label{tbl:alltechniques}
  \vspace{-0.3cm}
\end{table*}

\chgd{In summary, deep learning is becoming prevalent thanks to big data and fast computation, and software engineering is going through a new paradigm shift.} However, big data for deep learning has been relatively understudied, but is becoming critical in data-centric AI.
We cover the following topics in the next sections:

\begin{itemize}
    \item Data collection techniques for machine learning (Section~\ref{sec:datacollection}). 
    \item \chgd{Data validation, cleaning, and integration techniques for machine learning (Section~\ref{sec:datavalidationandcleaning}).}
    \item Robust training techniques for coping with noisy and poisoned data (Section~\ref{sec:robusttraining}).
    \item Fair training techniques for coping with biased data (Section~\ref{sec:fairtraining})
    \item \chgd{Overall findings and future directions (Section~\ref{sec:overall_findings})}.
\end{itemize}
\revision{We choose papers using three criteria. First, we include papers to cover the diverse areas in each section. Second, in each area, we select prominent papers preferably with many citations. Third, we cover recent techniques that are emerging, but are yet to be widely cited according to our judgment and tutorials.}

\chgd{
Note that Table~\ref{tbl:alltechniques} also specifies the data types each technique focuses on. For both robust and fair training (Sections~\ref{sec:robusttraining} and \ref{sec:fairtraining}), we mainly consider supervised learning. 
}

\vspace{-0.1cm}
\section{Data Collection}
\label{sec:datacollection}

\chgd{Our coverage of data collection originates from a survey\,\cite{DBLP:journals/tkde/RohHW19} by two of the authors, so we keep it brief with new updates based on a tutorial\,\cite{DBLP:journals/pvldb/Whang020}.} There are three main approaches for data collection. First, \emph{data acquisition} is the problem of discovering, augmenting, or generating new datasets. Second, \emph{data labeling} is the problem of adding informative annotations to data so that a machine learning model can learn from them. Since labeling is expensive, there is a variety of techniques to use including semi-supervised learning, crowdsourcing, and weak supervision. Finally, if one already has data, \emph{improving existing data and models} can be done instead of acquiring or labeling from scratch.


\vspace{-0.1cm}
\subsection{Data Acquisition}
\label{sec:dataacquisition}

If there is not enough data, the first option is to perform data acquisition, which is the process of finding datasets that are suitable for training machine learning models. In this survey we cover three approaches: data discovery, data augmentation, and data generation. \emph{Data discovery} is the problem of indexing and searching datasets. \chgd{\emph{Data augmentation} takes labeled examples and distorts or combines them to generate synthetic examples.} If there is not enough data around, the last resort is to take matters in one's own hands and create datasets using crowdsourcing or synthetic \emph{data generation} techniques.

\subsubsection{Data Discovery}

\emph{Data discovery} is the problem of indexing and searching datasets that exist either in corporate data lakes\,\cite{DBLP:conf/cidr/TerrizzanoSRC15,DBLP:conf/icde/FernandezAKYMS18} or the Web\,\cite{DBLP:journals/pvldb/CafarellaHLMYWW18}. \chgd{One example is the Goods system\,\cite{DBLP:conf/sigmod/HalevyKNOPRW16}, which searches tens of billions of datasets in Google's data lake. Goods takes a post-hoc approach where it crawls the datasets from multiple sources and extracts metadata to maintain a central dataset catalog, which does not require any work from the dataset owners.} Each entry in the catalog contains metadata about one dataset including its size, provenance on which job created it and which job read it, and schema information. Goods provides search, monitoring, and dataset annotations as well. \chgd{A public version of Goods called Google Dataset Search\,\cite{DBLP:conf/www/BrickleyBN19} supports science dataset searching. More recently, these data discovery tools have become more interactive. A representative system is Juneau\,\cite{DBLP:conf/sigmod/ZhangI20}, which is an interactive data search and management tool built on top of the Jupyter Notebook data science platform. Here the key technical challenge is finding the related tables. Juneau uses similarity measures for comparing records and schemas and provenance information that intuitively captures the purpose of creating each data set. Finding tables that can be joined or unioned in data lakes efficiently is critical, and LSH-based algorithms that perform set overlap search or unionable attribute retrieval on tables have been proposed\,\cite{DBLP:journals/debu/MillerNZCPA18}.}

\subsubsection{Data Augmentation}

For data augmentation, a popular method for generating data in the machine learning community is generative adversarial networks (GANs)\,\cite{DBLP:conf/nips/GoodfellowPMXWOCB14,DBLP:journals/corr/Goodfellow17,DBLP:journals/corr/abs-1807-04720}. \chgd{We start from a training set that has real data.} There are two components: a {\em generator} that generates fake data that is realistic using some random noise as an input and a {\em discriminator} that tries to distinguish the real data from the fake data of the generator. \chgd{The generator and discriminator are trained in an adversarial fashion.} One limitation of a GAN is that it cannot generate data that is completely different than the existing data. Using policies\,\cite{DBLP:conf/nips/RatnerEHDR17,DBLP:journals/corr/abs-2004-03264} is a way to complement that limitation where one can apply various custom transformations provided by domain experts as long as the data remains realistic. AutoAugment\,\cite{DBLP:conf/cvpr/CubukZMVL19} automates this process where the idea is to have a controller that suggests a strategy for applying transformations with certain probabilities and magnitudes on the data. The system then trains a child model on this augmented data and measures the accuracy on a validation set. This result is then used to decide whether the strategy produces useful data that is within realistic bounds and should thus be used.

The data augmentation literature continues to grow rapidly. Mixup\,\cite{DBLP:conf/iclr/ZhangCDL18,DBLP:conf/nips/BerthelotCGPOR19,DBLP:conf/iccv/YunHCOYC19,DBLP:conf/iclr/BerthelotCCKSZR20,DBLP:conf/iclr/HendrycksMCZGL20} has been proposed as a simple, but effective augmentation technique where the key idea is to mix pairs of data points of different classes. The additional data effectively regularizes the model to predict in-between training data points assuming linearity. Model patching\,\cite{modelpatching} utilizes GANs to augment the data of specific subgroups of a class so that the model accuracy is similar across subgroups.

\subsubsection{Data Generation}

\chgd{Another option for collecting or acquiring new data is to generate data. A popular option is to use crowdsourcing platforms like Amazon Mechanical Turk\,\cite{amazonmechanicalturk} where one can create tasks and pay human workers to create or find data. For example, a task may ask workers to find face images of a certain demographic from public websites\,\cite{DBLP:conf/sigmod/TaeW21}.} In addition, one can use a simulator or generator for specific domains, e.g., Hermoupolis~\,\cite{DBLP:conf/pkdd/PelekisNTST13} for mobility data and Crash to Not Crash~\,\cite{DBLP:conf/aaai/KimLHS19} for driving data. \chgd{Domain randomization\,\cite{DBLP:conf/iros/TobinFRSZA17,DBLP:conf/cvpr/TremblayPABJATC18} is an effective technique for generating a wide range of realistic data from a simulator by varying its parameters.} We note that GANs also generate new data, but they require sufficient amounts of real data for model training.

\subsection{Data Labeling}
\label{sec:datalabeling}

Once there are enough datasets, the next step is to label the examples. We cover data labeling techniques for utilizing existing labels and manually or automatically labeling from no labels. 

\subsubsection{Utilize Existing Labels}

\chgd{The traditional approach for labeling is semi-supervised learning\,\cite{DBLP:journals/kais/TrigueroGH15,zhu2008} where the idea is to use existing labels to predict the other labels. One can utilize existing machine learning benchmarks\,\cite{Dua:2019,kaggle} that provide labeled data for a variety of tasks.} The simplest form is {\em Self-training}\,\cite{Yarowsky:1995:UWS:981658.981684} where a model is trained on the available labeled data and then applied to the unlabeled data. Then the predictions with the highest confidence values are trusted and added to the training set. This approach assumes that we can trust the high confidence, but there are other techniques including Tri-training~\cite{Zhou:2005:TEU:1092713.1092809}, Co-learning~\cite{DBLP:conf/ictai/ZhouG04}, and Co-training~\cite{Blum:1998:CLU:279943.279962} that do not rely on this assumption.

\subsubsection{Manual Labeling from No Labels}

If there are no labels to start with, but one has funds to employ workers, a standard approach is to use crowdsourcing platforms like Amazon Mechanical Turk to perform labeling. Since labeling is such an important task, there are labeling-specialized services like Amazon Sagemaker Ground truth\,\cite{amazonsagemaker} and Google Cloud Labeling\,\cite{gcp}. When using Sagemaker, one can choose labeling tasks and recruit labelers who are assisted with a UI and tools to label the data. Sometimes, crowdsourcing may not be feasible because the workers do not have the right expertise. Hence, the last resort is to rely on domain experts, but this option can be expensive. 

Active learning\,\cite{DBLP:series/synthesis/2012Settles,DBLP:reference/sp/2015rsh} is an effective method to reduce the crowdsourcing cost. The idea is to ask human labelers to label uncertain examples that, when answered, are likely to improve model accuracy the most. While a full coverage of active learning is out of scope, the example selection techniques can largely be categorized into identifying uncertain examples and using decision theoretic approaches to analyze the effect of a newly-labeled example on the model accuracy.

\subsubsection{Automatic Labeling from No Labels}

Recently, {\em weak supervision} is becoming popular where the idea is to (semi-)automatically generate labels that are not perfect (therefore called ``weak'' labels), but at scale where the larger volume may compensate for the lower label quality. Weak supervision is useful in applications where there are few or no labels to start with. \chgd{Early techniques include crowdsourcing and distant supervision\,\cite{DBLP:conf/acl/MintzBSJ09}, which uses external knowledge bases to generate labels for the training data. More recently, data programming builds on these techniques where multiple labeling functions are developed and combined to generate weak labels.}

\chgd{Snorkel\,\cite{DBLP:journals/vldb/RatnerBEFWR20,Ratner:2017:SRT:3173074.3173077,DBLP:conf/sigmod/BachRLLSXSRHAKR19} is the seminal system for data programming. Given user-provided labeling functions (e.g., Python functions that detect spam), Snorkel combines them in one generative model by intuitively taking a probabilistic consensus.} Then given unlabeled data, Snorkel can generate probabilistic labels. The unlabeled data and the probabilistic labels are used to train a final discriminative model like a deep neural network. Another way to combine labeling functions is to use majority voting. \chgd{Empirically, the number of labeling functions determines whether a generative model or majority voting is better. Snuba\,\cite{DBLP:journals/pvldb/VarmaR18} automates the process of constructing labeling functions using a small labeled dataset, if that is available.}


\subsection{Improving Existing Data}
\label{sec:improvingexisting}

In addition to searching and labeling datasets, one can also improve the quality of existing data and models. This approach is useful in several scenarios. Suppose the target application is novel or non-trivial where there are no relevant datasets outside, or collecting more data no longer benefits the model's accuracy due to its low quality. \chgd{Here a better option may be to improve the existing data. One effective approach is to improve the labels through {\em re-labeling}.} Sheng et al.\,\cite{DBLP:conf/kdd/ShengPI08} demonstrates the importance of improving labels by showing the model accuracy trends against more training examples for datasets with different qualities. As the data quality decreases, even if more data is used, the accuracy of the model does not increase from some point and plateaus. In this case the only way to improve the model accuracy is to improve the label quality, which can be done by re-labeling and taking majority votes on multiple labels per example. \chgd{In fact, one could clean the entire data including labels, which naturally leads to the next section where we cover data validation, cleaning, and integration.}

\section{\chgd{Data Validation, Cleaning, and Integration}}
\label{sec:datavalidationandcleaning}

It is common for the training data to contain various errors. Machine learning platforms like TensorFlow Extended (TFX)\,\cite{DBLP:conf/kdd/BaylorBCFFHHIJK17} have data validation\,\cite{Polyzotis:2017:DMC:3035918.3054782} components to detect such data errors in advance using data visualization and schema generation techniques. Data cleaning can be used to actually fix the data, and there is a heavy literature\,\cite{DBLP:books/acm/IlyasC19} on various integrity constraints. \chgd{However, recent studies\,\cite{cleanmlli,10.1145/3506712} show that cleaning the data before machine learning by only fixing well-defined errors does not necessarily benefit machine learning accuracy. Instead, it is more effective to clean {\em for machine learning} with the direct purpose of improving accuracy\,\cite{DBLP:journals/debu/NeutatzCA021} and making the model training more robust to noise in the data\,\cite{DBLP:conf/icml/LiuPRT21}. A recent survey\,\cite{10.1145/3506712} mentions that robust training is considered more effective than data cleaning before model training. Data noise can also be adversarial where it contains malicious poisoning, and cleaning against this is called data sanitization in the security community. Yet another issue is incorporating AI ethics like model fairness\,\cite{DBLP:journals/debu/BiessmannGRL021} where data may be biased, which may cause the trained model to be discriminatory. So far the data validation literature does not cover robust and fair training in depth, but these areas are heavily studied in the machine learning community, so we make a connection by elaborating on their techniques in Sections~\ref{sec:robusttraining} and \ref{sec:fairtraining}, respectively. }

\subsection{Data Validation}
\label{sec:datavalidation}

\chgd{Data visualization is a widely-used and effective way to validate data for machine learning (see a tutorial\,\cite{Polyzotis:2017:DMC:3035918.3054782} and survey\,\cite{DBLP:journals/sigmod/PolyzotisRWZ18}). Compared to traditional data cleaning, visualization is effective for a human to perform quick, but important sanity checks on the data to prevent larger errors downstream. A representative open-source tool is Facets\,\cite{facets}, which shows various statistics and the contents of datasets that can be used for sanity checks on data to prevent larger errors downstream.} In addition to manual visualization, there has also been research on {\em automatic generation} of new visualizations\,\cite{DBLP:journals/debu/LeeP18} that can be used for validation purposes. SeeDB\,\cite{DBLP:journals/pvldb/VartakRMPP15} is a seminal framework that repeatedly generates visualizations of interest. To capture the notion of interestingness, SeeDB uses a deviation-based utility metric that gives a high value when groupings of the data result in different probability distributions. 

Automatically generating visualizations can run into the problem of false positives, so there is also a line of research that proposes {\em false discovery control} techniques. 
\chgd{CUDE\,\cite{DBLP:conf/sigmod/ZhaoSZBUK17} controls false discovery in the context of multiple hypothesis testing for visual interactive data exploration. Here users can repeatedly generate visualizations and mark the ones that are significant. Based on this user feedback, the goal is to automatically choose the visualizations that are significantly interesting in a statistical sense.}

\begin{figure}[h]
  \includegraphics[width=\columnwidth]{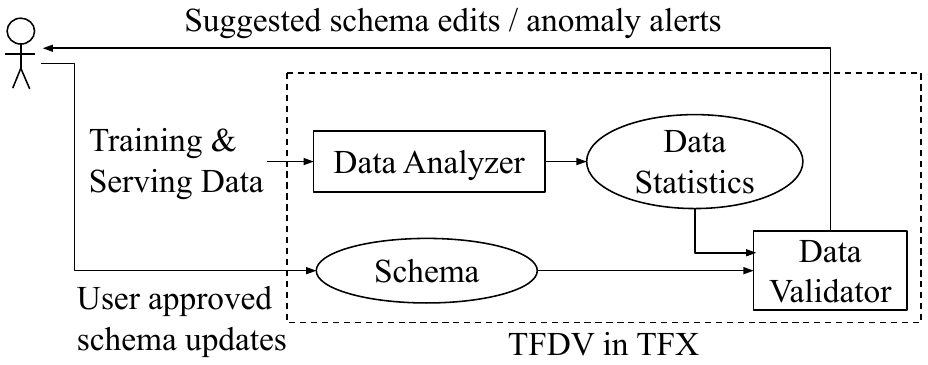}
  \caption{TensorFlow Data Validation (TFDV)\,\cite{breck2019data} uses a user-approved schema to validate the statistics of the training and serving data.}
  \label{fig:tfdv}
\end{figure}

{\em Schema-based validation}\,\cite{DBLP:conf/kdd/BaylorBCFFHHIJK17,Polyzotis:2017:DMC:3035918.3054782} is widely used in practice. \chgd{Tensorflow Data Validation (TFDV)\,\cite{breck2019data,DBLP:journals/debu/DrevesHPPRC21} assumes a continuous training setting where input data periodically streams in as shown in Figure~\ref{fig:tfdv}.} TFDV generates a data schema from previous data sets and uses the schema to validate future data sets and alert users on data anomalies. For each anomaly, TFDV provides concrete action items to possibly fix the root cause. A schema here is different from a traditional database schema where it contains a summary of data statistics of the features. In case a new dataset violates the current schema, either the data needs to be fixed, or the schema needs to be updated, and the user makes the decision.

\chgd{More recently, data validation systems are equipped with additional functionalities. Deequ\,\cite{DBLP:journals/pvldb/SchelterLSCBG18,DBLP:conf/icde/SchelterGSRKTBL19} allows one to write data quality constraints declaratively, which then are automatically generated into unit tests. The mlinspect library\,\cite{DBLP:conf/cidr/GrafbergerSS21} enables declarative machine learning pipeline inspection. Other additions include automatic identification of error types\,\cite{DBLP:conf/edbt/RedyukKMS21}, testing the impact of errors on models\,\cite{DBLP:conf/edbt/SchelterRB21}, ease of usage\,\cite{DBLP:conf/cidr/MelgarDGGHJKL0L21}, and efficient human-in-the-loop validation\,\cite{DBLP:journals/debu/XinPTWGHJP21}.}

\subsection{Data Cleaning}
\label{sec:datacleaning}

{\em Data cleaning} has a long history of removing various well-defined errors by satisfying integrity constraints including key constraints, domain constraints, referential integrity constraints, and functional dependencies. For an introduction see the book {\em Data Cleaning}\,\cite{DBLP:books/acm/IlyasC19}. \chgd{There is also a recent survey on data cleaning techniques for machine learning and vice versa\,\cite{10.1145/3506712}.}

We first introduce one of the state-of-the-art data cleaning techniques to give a sense of how sophisticated these techniques have become. HoloClean\,\cite{DBLP:journals/pvldb/RekatsinasCIR17} repairs data using probabilistic inference using three main ingredients: satisfying various integrity constraints, using external dictionaries to check the validity of values, and using quantitative statistics. 

Unfortunately, only focusing on fixing the data does not necessarily guarantee the best model accuracy. At first glance, it seems that perfectly cleaning the data would be most useful for the model training. However, the notion of clean data is not always clear cut, and removing all possible errors is not always feasible. CleanML\,\cite{cleanmlli} is a framework that evaluates various data cleaning techniques and seeing if they actually help model accuracy. The authors show that data cleaning does not necessarily improve downstream machine learning models. In fact, the cleaning may sometimes have a negative effect on the models. However, by selecting an appropriate machine learning model, one can eliminate the negative effects of data cleaning. Also there is no single cleaning algorithm that performs the best, and one must adaptively choose the algorithm depending on the type of noise to clean. \chgd{Moreover, many data cleaning primitives have high-impact parameters like thresholds that need to be tuned, similar to machine learning hyperparameter tuning.} Hence, data cleaning techniques that are not originally designed for machine learning must be used carefully.

\begin{figure}[h]
  \includegraphics[width=\columnwidth]{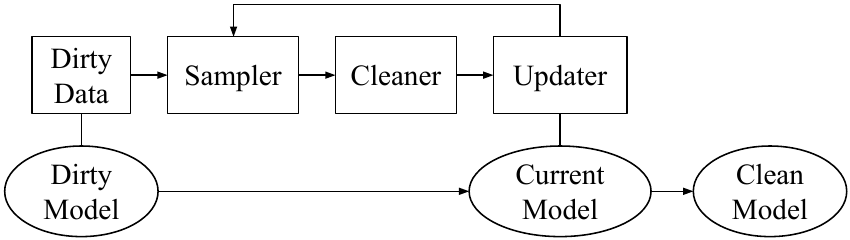}
  \caption{ActiveClean\,\cite{DBLP:journals/pvldb/KrishnanWWFG16} iteratively selects data that is likely to be dirty and cleans it.}
  \label{fig:activeclean}
\end{figure}

Recently there are data cleaning techniques with the specific purpose of improving model accuracy\,\cite{DBLP:conf/kdd/DongR19}. ActiveClean\,\cite{DBLP:journals/pvldb/KrishnanWWFG16} is a seminal framework that iteratively cleans samples of dirty data and updates the model. Figure~\ref{fig:activeclean} shows the workflow where there is a sampler that chooses an example that is likely to be dirty, and data quality rules can be used to identify such dirty samples. The reason for sampling data is that cleaning the entire data is presumed to be very expensive. Each sample can be cleaned by an oracle or domain expert. \chgd{Then the model is updated to be more accurate and also chooses the next sample.} ActiveClean has theoretical guarantees where, by repeatedly training a model on the clean sample plus previously-cleaned data, the model eventually obtains an accuracy as if it was trained on clean data only. ActiveClean assumes convex loss models like SVMs, and the data cleaning is assumed to be done perfectly. 

Another branch of research is to clean the labels for the purpose of improving model accuracy. TARS\,\cite{DBLP:journals/pvldb/DolatshahTWP18} is a system that predicts model accuracy out of noisy labels that are produced from crowdsourcing. \chgd{TARS first chooses labels that are likely to be flipped because they were labeled by poor-performing workers and thus have low confidence values.} TARS then estimates how much the model will improve if the label is flipped after cleaning. The confidence values of labels can be computed by using confusion matrices of workers, which capture the history of how well they performed in past tasks. A confusion matrix thus contains the previous false positive, false negative, true positive, and true negative rates. Given the probability that a label is flipped, TARS estimates the resulting model accuracy and subtracts that by the estimated accuracy of the current model to determine whether the label is worth examining. Hence, TARS can selectively clean labels that are expected to benefit model accuracy the most.

\chgd{More recently, there are more systematic approaches to support data cleaning for machine learning. One study\,\cite{DBLP:journals/debu/RenggliRGK0021} shows how data quality issues affect MLOps and proposes various solutions to tackle them. For example, CPClean\,\cite{DBLP:journals/pvldb/KarlasLWGC0020} is proposed to analyze how missing data impacts the certainty of predictions. Another work\,\cite{DBLP:journals/debu/NeutatzCA021} distinguishes data cleaning {\em before} machine learning versus {\em for} machine learning and suggests to clean data throughout the entire machine learning pipeline. Some common challenges include handling multimodal data and data that changes over time. }

\subsection{Data Sanitization}
\label{sec:datasanitization}

{\em Data poisoning} has recently become a serious issue because changing a fraction of training data, which may come from an untrusted source, may alter the model's behavior. Compared to dirty data, there is a malicious intent to make the model fail. Data poisoning is a real problem because data is now easier to publish through dataset search engines. A dataset owner can simply post metadata to the public, which will be automatically crawled by the search engine. Then one can simply harvest that data using web crawlers without knowing that the data is poisoned. Data sanitization\,\cite{DBLP:conf/sp/CretuSLSK08} is the problem of defending against such poisoning attacks and can be viewed as an extreme version of data cleaning.

\chgd{A simple type of data poisoning is called label flipping where a label of a training example is flipped from one class to another, but other works generate new data as well. Recently, data poisoning techniques have become much more sophisticated and therefore harder to defend against\,\cite{DBLP:conf/nips/ShafahiHNSSDG18,DBLP:conf/icml/ZhuHLTSG19}.} We illustrate a state-of-the-art data poisoning techniques for deep learning\,\cite{DBLP:conf/icml/ZhuHLTSG19}. A major challenge when poisoning data for deep learning is that the victim's model cannot be easily analyzed. Hence, transferable poisoning attacks have been proposed, which can still succeed without accessing the victim's model. The idea is to train an ensemble of substitute models, which are assumed to be similar to the victim's model. Any attack that negatively affects the substitute models will presumably attack the victim's model as well. 
Given a set of clean data points of different classes, the poisoning algorithm adjusts the clean points to ``move closer'' to the target within the feature space and form a convex polytope that surrounds it to maximize the chances of the target to be misclassified.

\begin{figure}[h]
  \includegraphics[width=\columnwidth]{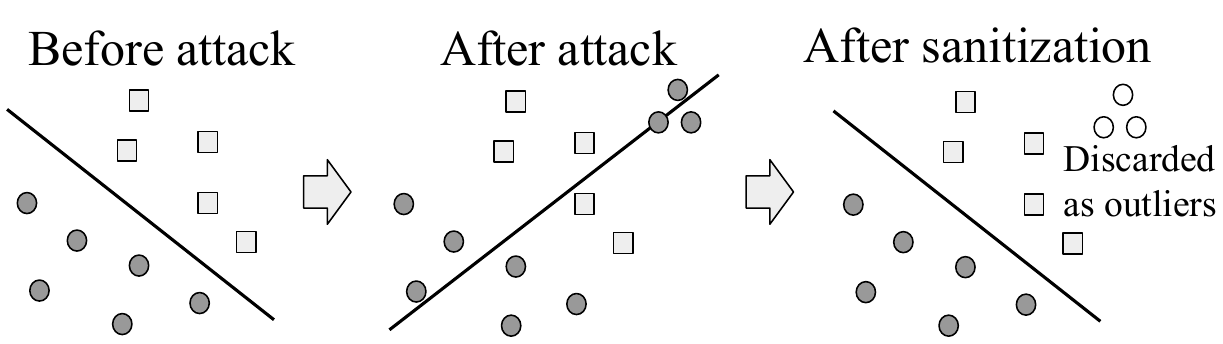}
  \caption{Data Sanitization\,\cite{DBLP:conf/sp/CretuSLSK08} identifies and discards data poisoning for better model accuracy.}
  \label{fig:datasanitization}
\end{figure}

How do we defend against such data poisoning using data sanitization? \chgd{The main approach is to perform outlier detection to detect poisonings and discard them.} Figure~\ref{fig:datasanitization}  shows a simple setting where a classifier's behavior changes after introducing poisoning (top right data points). \chgd{If the data sanitization can identify and discard these points as outliers, then the model's accuracy can be restored. Compared to traditional outlier detection, the challenge is that poisonings are intentionally designed by the adversary to be difficult to detect while reducing model accuracy.} Data sanitization techniques\,\cite{DBLP:conf/sp/CretuSLSK08,DBLP:journals/air/HodgeA04,DBLP:journals/corr/abs-1802-03041} have been proposed throughout the years, and a recent study\,\cite{DBLP:journals/corr/abs-1811-00741} evaluates various defenses by developing attacks and seeing if the defenses work are still effective. Unfortunately, the conclusion is that no technique can adequately defend against carefully designed attacks. We suspect that data poisoning and sanitization techniques will continue to evolve and compete with each other. 

\chgd{\subsection{Multimodal Data Integration}
\label{sec:dataintegration}

Another dimension of data management to consider is the issue of multimodal data integration\,\cite{DBLP:journals/pami/BaltrusaitisAM19}. So far, we implicitly assumed single-source datasets, but in practice, data scientists often deal with multimodal data from multiple sources. For example, autonomous vehicles can generate a wide range of data including multiple video streams, radar and lidar data, and thousands of irregular times series from the Controller Area Network (CAN) of the vehicle. Analyzing all of this data together requires some form of data integration. In machine learning, two relevant integration techniques are alignment and co-learning. Alignment is to find relationships of sub-components of instances that have multiple modalities. For example, if there are multi-view time series, one can perform subsampling, forward or backward filling, or aggregate in time windows so that the time series can be better integrated. Co-learning is to train better on a modality using a different modality. For example, if there are embeddings from different modalities, one approach is to concatenate them together for a multimodal representation. In general, data integration is by itself a large research area that has been studied for decades\,\cite{DBLP:journals/debu/StonebrakerI18,DBLP:books/daglib/0029346}, although not all techniques are relevant to machine learning.}

\section{Robust Model Training}
\label{sec:robusttraining}

Even after collecting the right data and cleaning it, data quality may still be an issue during model training. It is widely agreed that real-world datasets are dirty and erroneous despite the data cleaning process. \chgd{As summarized in Table \ref{table:poisoning_type}, these flaws in datasets can be categorized depending on whether data values are noisy or missing and depending on whether these flaws exist in data features\,(attributes) or labels.}


\chgd{The problem of data poisoning has been studied in theory\,(i.e., robust statistics) and practice for over fifty years and has gained a lot of attention in the machine learning community\,\cite{tukey1960survey, huber1992robust}. It starts with a basic question, `can the machine learning model learn and predict as if the data was not corrupted?' and aims to develop machine learning algorithms robust to the worst-case corruptions where we cannot recover the entire clean information from the data. It mainly considers the corruptions in data features, 
which include outliers and adversarial examples. 

Statistical approaches like robust mean estimation\,\cite{liu2021robust} aims to recover the mean of the distribution in the presence of data flaws. Convex programming\,\cite{diakonikolas2019robust} and filtering\,\cite{cheng2019high} address the problems by assigning a score to each data point based on the degree to which the sample is considered corrupted. This series of studies have been inspiring a lot of machine learning robustness optimization techniques such as loss reweighting and sample selection.
In addition, robust machine learning involves many problems depending on what sorts of damages we consider. For example, privacy machine learning aims to respect the privacy of the users providing the data\,\cite{qayyum2020secure}. 
}

{\footnotesize
\begin{table}[t!]
\begin{tabular}{|c|c|c|} \hline
         & Noisy & Missing \\\hline 
\!\!Features\!\! & \!\!\makecell{Adversarial Learning\\(Section \ref{sec:noisy_feature})}\!\! & \makecell{Data Imputation\\(Section \ref{sec:missing_feature})} \\\hline
Labels           & \makecell{Robust Learning\\(Section \ref{sec:noisy_label})}              & \!\!\makecell{Semi-Supervised Learning\\(Section \ref{sec:missing_label})}\!\! \\\hline 
\end{tabular}
\caption{\chgd{Types of data poisoning covered in this survey.}}
\label{table:poisoning_type}
\end{table}
}

\subsection{Noisy Features}
\label{sec:noisy_feature}

Noisy features are often introduced by adversarial attacks. Among several types of attacks, we focus on the \emph{poisoning attack}, which is known as contamination of the training data, to be aligned with the main theme of this survey. During the training phase of a machine learning model, an adversary tries to poison the training data by injecting maliciously designed data to deceive the training procedure.
\chgd{Besides the adversarial noise, noisy features can include natural noise like image blurring and color noise, possibly not removed by data cleaning. There have been some approaches to successfully denoising the natural noise using Sparse Coding\,\cite{shang2008denoising} and Feature Attention\,\cite{anwar2019real}, but they are out of the scope of this survey.} 

Either features or labels or both can be the target of the poisoning attack. The poisoning attack can be done in three ways depending on the capability of adversaries. 
First, an adversary can randomly perturb the labels, i.e., by assigning other incorrect labels, picked from a random distribution, to a subset of training data. \chgd{Since the label flipping result in overfitting to wrong labels like noisy labels, the robust training methods for this type of attack will be discussed in Section \ref{sec:noisy_label}.}
Second, an adversary is more powerful and can corrupt the features of the examples possibly determined by analyzing the training algorithm\,\cite{DBLP:conf/pkdd/BiggioCMNSLGR13}. \chgd{The corrupted features deceive the model into making wrong predictions.}
\chgd{Third, {unlike manipulating the features, an adversary may add adversarial examples into the training data such as out-of-distribution examples. These examples lead to a sharp drop in generalization capability of machine learning models under distributional shifts.}}
For more details, the reader can refer to \cite{DBLP:journals/corr/abs-1810-00069, shen2021towards}.

Various defense strategies have been actively studied for robust training on adversarial examples\,(e.g., noisy features). Most of the current strategies are not adaptive to all types of attacks, but are effective to only a specific type. We summarize a few well-known, representative strategies in this section.

\chgd{Most notably, in \emph{adversarial training},
the robustness of a model can be improved using a modified objective function based on the fast gradient sign method\,\cite{DBLP:journals/corr/GoodfellowSS14}. It is defined as a weighted sum of an usual loss function on clean examples and the loss function on adversarial examples.}
\chgd{By this regularization, the model is forced to predict the same class for legitimate and perturbed examples in the same direction.} 

Knowledge distillation has been shown to be effective for adversarial attacks\,\cite{DBLP:conf/sp/PapernotM0JS16}. \emph{Defensive distillation} is almost the same as typical knowledge distillation, except that the same network architecture is used for both the original network and the distilled network. Specifically, instead of hard labels, where only the true label has the probability 1 in a probability vector, \emph{soft targets}, which are generated by the original network as the prediction result, are used for training the distilled network. The benefit of using soft targets comes from the additional knowledge found in probability vectors compared to hard class labels\,\cite{DBLP:conf/sp/PapernotM0JS16}.

\emph{Feature squeezing}\,\cite{DBLP:conf/ndss/Xu0Q18}  reduces the degree of freedom
available to an adversary by squeezing out unnecessary input features. If the original and squeezed inputs result in substantially different outputs by a model, the corresponding input is determined to be adversarial. A popular squeezing technique for images is reducing the color depth on a pixel level.

Another idea is to detect adversarial examples using separate classification networks\,\cite{DBLP:conf/iclr/MetzenGFB17}. A sub-network, called an \emph{adversary detection network} or simply a detector, is trained to produce an output that indicates the probability of the input being adversarial. For this purpose, a classification network is trained using only non-adversarial examples, and adversarial examples are generated for each example in the training set. \chgd{Then, the detector is trained using both the original dataset and the corresponding adversarial dataset. MagNet\,\cite{DBLP:conf/ccs/MengC17} falls into this category, and it also contains a reformer that corrects an adversarial example to a similar legitimate example using autoencoders.}

\subsection{Missing Features}
\label{sec:missing_feature}

Since missing data can reduce the statistical power and produce biased estimates, data imputation has been an active research topic in statistics and machine learning.
\chgd{Missing features can occur in any types of data, but, in this paper, we focus on \emph{multivariate time-series} data because 
its high input rate and sensor malfunction generate missing values very often.} 

Missing values in multivariate time-series data are ubiquitous in many practical applications ranging from healthcare, geoscience, astronomy, to biology and others. They often inevitably carry missing observations due to various reasons, such as medical events, saving costs, anomalies, inconvenience, and so on. These missing values are usually informative where the missing value and patterns provide rich information about target labels in supervised learning tasks. 

\begin{figure}[h]
  \includegraphics[width=\columnwidth]{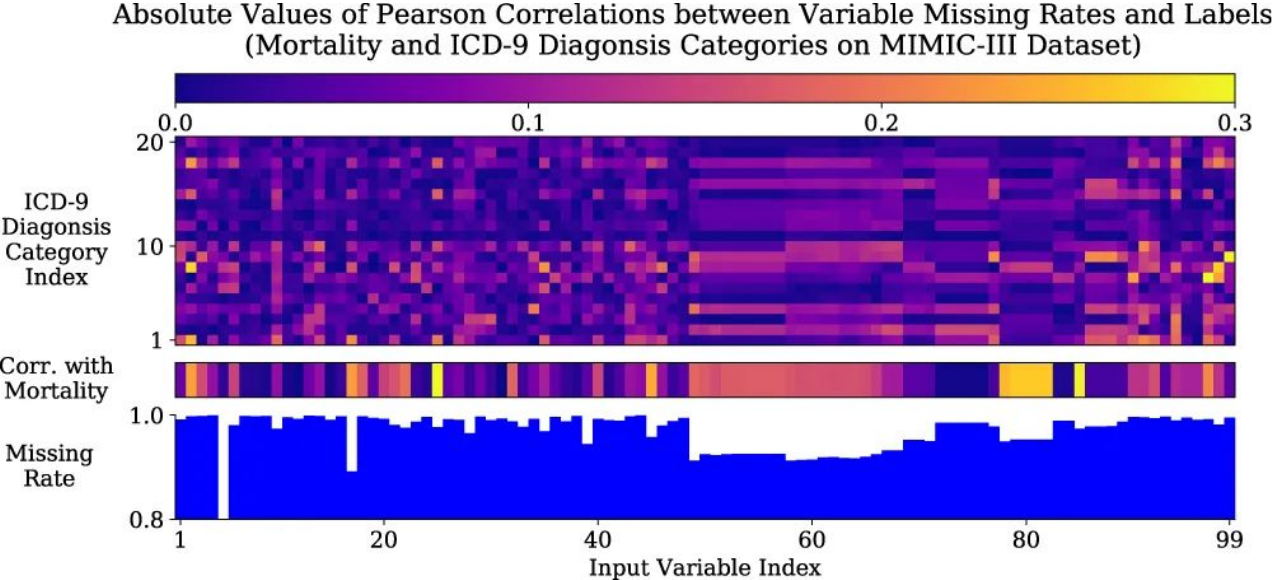}
  \caption{Informative missingness in the MIMIC-III dataset\,\cite{CheEtAl_nature_sr18}.}
  \label{fig:informativemissingness}
\end{figure}

We first describe informative missingness. Figure~\ref{fig:informativemissingness} shows MIMIC-III critical care dataset\,\cite{CheEtAl_nature_sr18}. Starting form the bottom, there are the missing rate of each variable, the correlation between missing rate of each variable and mortality, and the correlation between missing rate of each variable and each ICD-9 diagnosis category. Here, we observe that the values of missing rates are correlated with labels, where the values with low missing rates are highly correlated with the labels. In other words, the missing rate of variables of each patient is useful, and this information is more useful for the variables that are more often observed in the dataset.

\begin{figure*}[t]
  \centering
  \includegraphics[width=0.8\linewidth]{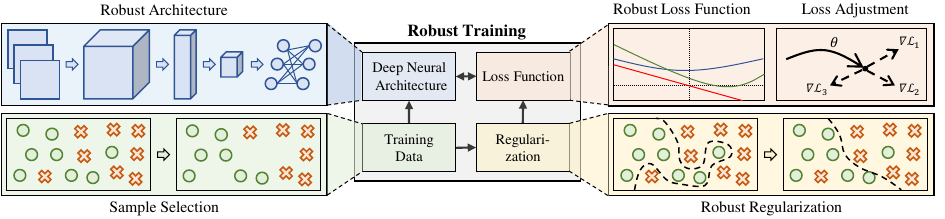}
  \vspace*{-0.2cm}
  \caption{Robust training categorization\,\cite{DBLP:conf/icml/SongK019}.}
  \label{fig:robusttraining}
  \vspace{-0.3cm}
\end{figure*}

For existing approaches, a simple solution is to omit the missing data and to perform analysis only on the observed data, but it does not provide good performance when the missing rate is high and the samples are inadequate. Another solution is to fill in the missing values with substituted values, which is known as \emph{data imputation}. However, these methods do not capture variable correlations and may not capture complex patterns to perform imputation. Combining the imputation methods with prediction models often results in a two-step process where imputation and prediction models are separated; the missing patterns are not effectively explored in the prediction model, thus leading to suboptimal analysis results.

GRU-D\,\cite{CheEtAl_nature_sr18} is a deep learning model based on the gated recurrent unit\,(GRU) to effectively exploit two representations of informative missingness patterns---masking and time interval. Masking informs the model of which inputs are observed or missing, while time interval encapsulates the input observation patterns. GRU-D captures the observations and their dependencies by applying masking and time interval, which are implemented using a decay term, to the inputs and network states of the GRU, and jointly train all model components through back-propagation. GRU-D not only captures the long-term temporal dependencies of time-series observations, but also utilizes the missing patterns to improve the prediction results.

We elaborate on the two components of GRU-D: \emph{making} and \emph{time interval}. 
\chgd{The value of a missing variable tends to be close to some default value if its last observation happened a long time ago, because 
the influence of the last observation fades away over time.} 
As lots of missing patterns are informative and potentially useful in prediction tasks, but unknown and possibly complex, the goal is to learn decay rates from the training data rather than fixing them a priori. The GRU-D model incorporates two different trainable decay mechanisms. For a missing variable, an input decay $\gamma_x$ is added to decay it over time toward the empirical mean, instead of using the last observation as it is. A hidden state decay $\gamma_h$ in GRU-D has an effect of decaying the extracted features (GRU hidden states) rather than raw input variables directly.


\chgd{We now extend the discussion to cover tabular data and present interesting studies in statistics, machine learning, and query optimization. 

\begin{itemize}
\item {\em Statistics}: MICE\,\cite{JSSv045i03}, which is one of the most commonly-used packages in R, creates multiple imputed datasets to take care of uncertainty in missing values. By default, linear regression is applied to predict missing values. Besides, users can build models on all imputed datasets for evaluation and combine the results from these models to obtain a consolidated output. 
\item {\em Machine learning}: XGBoost\,\cite{ChenG16} internally handles missing values. It implements gradient boosted decision trees, and node splits are determined by considering missing values. In more detail, when a value is missing, the instance is classified into the default direction because there is nothing to evaluate for the split criteria. Here, the optimal default directions are learned from the data.
\item {\em Query optimization}: ImputeDB\,\cite{CambroneroFSM17} selectively applies imputation to a subset of records dynamically during query execution. The rationale behind this optimization is that the subset which imputation is needed for a \emph{specific} query is generally much smaller than the entire database. Thus, the computation spent for imputation is significantly saved. 
\vspace{-0.2cm}
\end{itemize}}

\subsection{Noisy Labels}
\label{sec:noisy_label}

\chgd{Because data labeling is done manually in many cases, incorrect or missing labels are, in fact, very common; the proportion of incorrect labels is reported to be 8--38\% in several real-world datasets\,\cite{DBLP:journals/corr/abs-2007-08199}.} 
%
\chgd{As an example of ANIMAL-10N\,\cite{DBLP:conf/icml/SongK019}, which is real-world noisy data with human annotation errors, human annotators mistakenly classified the Cheetah images as other animals like Jaguars instead of Cheetahs. In this example, it may be difficult to distinguish the patterns of Cheetahs and Jaguars, resulting in noisy labels in training data. So wrong annotations can be caused by such human errors. Similarly, labeling errors occur with data types other than images. For sentiment analysis, annotators often disagree on the polarity (e.g., positive or negative) of the sentiment expressed in the text\,\cite{wang2019learning}. 
%
Another type of error is software error. If there are many images to annotate, one may use automatic object recognition software. However, the object recognition itself may have errors.
Thus, many deep learning techniques have been developed to consider the existence of label noises and errors, which are more critical in deep learning than in conventional machine learning as a deep neural network completely memorizes such noises and errors because of its high expressive power.}


We explain what kinds of problems occur with noisy labels. In standard supervised learning, training data consists of example and label pairs $\{(x_i, y_i)\}_{i=1}^N$. In a practical setting, however, the label $y_i$ is actually $\tilde{y_i}$, which means it can be incorrect. 
If one trains powerful models like VGG-19 on noisy data, the model may simply memorize the noise as well and perform worse on clean data. The goal of the noisy label problem is to train the network as if there are no noisy labels.

Figure~\ref{fig:robusttraining} from a recent survey\,\cite{DBLP:conf/icml/SongK019} shows the categorization of robust training techniques. There are largely four components in the training procedure: deep neural architecture, loss function, input training data, and regularization. For each component, there are relevant robust training techniques. For deep neural architectures, robust architectures have been developed. For training data, various sample selection techniques have been proposed. For the loss function, robust loss functions and loss adjustment techniques have been proposed. More specifically, loss adjustment can be further divided into loss correction\,\cite{DBLP:conf/cvpr/PatriniRMNQ17}, loss reweighting\,\cite{DBLP:conf/nips/ChangLM17}, and label refurbishment\,\cite{DBLP:journals/corr/ReedLASER14}. For regularization, robust regularization techniques have been proposed. 

\begin{figure}[t!]
  \centering
  \includegraphics[width=0.85\columnwidth]{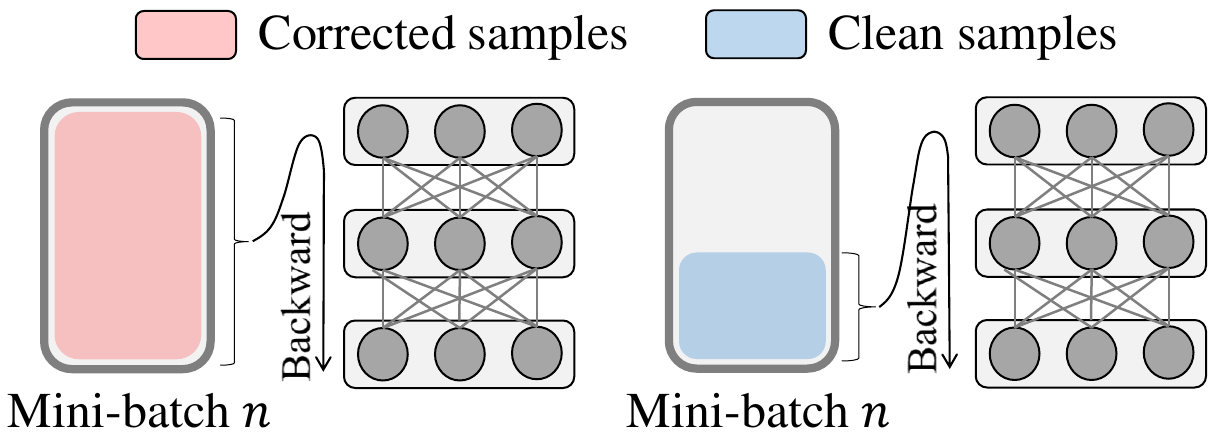} \\
  {\small (a) Loss correction. \hspace*{0.8cm} (b) Sample selection.}
  \vspace{-0.1cm}
  \caption{Two directions of robust training covered in this survey.}
  \label{fig:twodirections}
  \vspace{-0.3cm}
\end{figure}

In this survey, we focus on the most representative techniques: sample selection and loss correction techniques as illustrated in Figure~\ref{fig:twodirections}. \emph{Loss correction} is to correct the loss of \emph{all} samples before a backward step. The representative techniques include Bootstrap\,\cite{DBLP:journals/corr/ReedLASER14}, F-correction\,\cite{DBLP:conf/cvpr/PatriniRMNQ17}, and ActiveBias\,\cite{DBLP:conf/nips/ChangLM17}. \emph{Sample selection} is to select \emph{expectedly clean} samples to update the network. The representative techniques include Decouple\,\cite{DBLP:conf/nips/MalachS17}, MentorNet\,\cite{DBLP:conf/icml/JiangZLLF18}, and Coteaching\,\cite{DBLP:conf/nips/HanYYNXHTS18}.

We first introduce ActiveBias\,\cite{DBLP:conf/nips/ChangLM17}, which is a loss correction technique. ActiveBias performs a forward step on a given mini-batch and computes the sample importance for each sample. There are many statistics for the importance, e.g., variance of predictions. ActiveBias then corrects the loss by multiplying the normalized importance. If a label is noisy, then its importance $\mu_i$ decreases. The corrected loss is used to update the network.
We then explain sampling selection through its representative technique Coteaching\,\cite{DBLP:conf/nips/HanYYNXHTS18}, where the noise rate $\tau$ is assumed to be given. It then performs a forward step on a given mini-batch and selects the ($100-\tau$)\% low-loss samples as clean samples. The network is updated using the loss of the clean samples.

Although these two methods have improved the robustness to noisy labels, there are limitations of the two methods. Loss correction suffers from accumulated noise due to the large number of false corrections. Since all the examples are used for the training step, false corrections can accumulate for heavily noisy data. On the other hand, sample selection uses only clean samples having low losses and easy to classify. Hence, we may end up ignoring many useful, but hard samples that are classified as unclean.

\begin{figure}[t!]
  \centering
  \includegraphics[width=0.66\columnwidth]{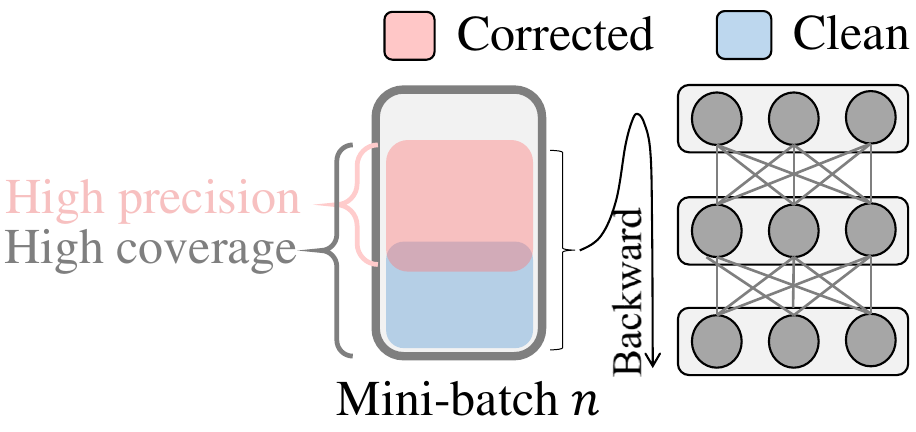}
  \vspace{-0.1cm}
  \caption{SELFIE is a hybrid of loss correction and sample selection\,\cite{DBLP:conf/icml/SongK019}.}
  \label{fig:selfie}
  \vspace{-0.3cm}
\end{figure}

SELFIE\,\cite{DBLP:conf/icml/SongK019} was proposed to overcome the above limitations by using a hybrid of loss correction and sample selection (see Figure~\ref{fig:selfie}). SELFIE introduces \emph{refurbishable} samples where labels can be corrected with high precision. The key issue of SELFIE is constructing the refurbishable and clean samples. For the clean samples, SELFIE adopts the small-loss trick\,\cite{DBLP:conf/nips/HanYYNXHTS18} and uses the ($100-\tau$)\% low-loss samples in the mini-batch. The refurbishable samples are ones that have consistent label predictions. Each label is replaced with the most frequently predicted label during the training step. For example, if an image is predicted mostly as a dog and only sometimes a cat, then the label predictions are considered consistent, and such a cat label is considered noisy and corrected to a dog. Finally, the loss is calculated for the refurbishable samples with correct labels and the clean samples; the samples that are neither refurbishable nor clean are discarded. The advantage of SELFIE is that it minimizes false corrections during the model training by selectively correcting refurbishable samples. As a result, the correction error of refurbishable samples is low. Also as the training progresses, the number of refurbishable samples also increases, so most training samples are exploited in the end.

Prestopping is another technique\,\cite{DBLP:conf/icml/SongK019} for avoiding overfitting to noisy labels by early stopping the training of a deep neural network before the noisy labels are severely memorized. The algorithm resumes training the early-stopped network using a maximal safe set, which maintains a collection of almost certainly true-labeled samples. 
MORPH\,\cite{DBLP:conf/kdd/SongKPS021} further improves Prestopping through a novel concept of \emph{self-transitional learning}, which automatically switches its learning phase at the transition point. The optimal transition point is determined without any supervision such as a true noise rate and a clean validation set, which are usually hard to acquire in real-world scenarios. MORPH rather estimates the noise rate by fitting the loss distribution to a one-dimensional and two-component Gaussian mixture model\,(GMM).

DivideMix\,\cite{DBLP:conf/iclr/LiSH20} is a recent technique that trains two networks simultaneously. At each epoch, a network models its per-sample loss distribution with a GMM to divide the dataset into a labeled set (mostly clean) and an unlabeled set (mostly noisy), which is then used as training data for the other network (i.e., co-divide). At each mini-batch, a network performs semi-supervised training using an improved MixMatch\,\cite{DBLP:conf/nips/BerthelotCGPOR19} method, which we cover in the next section. When training on the CIFAR-10 dataset with 40\% asymmetric noise, standard training with cross-entropy loss causes the model to overfit and produce over-confident predictions, making the loss difficult to be modeled by the GMM. Also, adding a confidence penalty during the warm up leads to more evenly-distributed loss. Finally, training with DivideMix can effectively reduce the loss for clean samples while keeping the loss larger for most noisy samples.

\subsection{Missing Labels}
\label{sec:missing_label}


We cover the issue of missing labels where training labels may not exist for either some or all examples. There are largely semi-supervised and unsupervised approaches. In semi-supervised approaches, clean labeled data exists together with unlabeled\,(or incorrectly labeled) data. The goal is to exploit unlabeled data to improve accuracy as much as possible. Here, the loss is defined as the supervised loss for labeled data plus the unsupervised loss for unlabeled data. The representative techniques include unsupervised loss\,(e.g., consistency loss) like Mean-Teacher\,\cite{DBLP:conf/nips/TarvainenV17} and augmentation techniques like MixMatch\,\cite{DBLP:conf/nips/BerthelotCGPOR19}. For unsupervised approaches, the representative techniques include self-supervised learning and generative models, and we will cover a self-supervised learning technique called JigsawNet\,\cite{DBLP:conf/eccv/NorooziF16}. 

\chgd{In Mean-Teacher\,\cite{DBLP:conf/nips/TarvainenV17}, the teacher model is the average of consecutive student models. Both the student and teacher models evaluate the input in a training batch.} The softmax output of the student model is compared with the one-hot label using a classification cost. Additionally, the output is compared with the teacher output using the consistency loss. After the weights of the student models are updated via gradient descent, the teacher model weights are updated as an exponential moving average of the student model weights. A training step with \emph{unlabeled} examples is done without the classification cost.

\chgd{In MixMatch\,\cite{DBLP:conf/nips/BerthelotCGPOR19}, to exploit an unlabeled dataset, it performs label guessing where stochastic data augmentation is applied to an unlabeled image $K$ times; then, each augmented image is fed through the classifier.} The average of these $K$ predictions is sharpened by adjusting the distribution's temperatures. The MixMatch algorithm mixes both labeled examples and unlabeled example with label guesses. In more detail, when mixing two images, the images are overlayed, and the labels are averaged, following the MixUp augmentation\,\cite{DBLP:conf/iclr/ZhangCDL18}.

\begin{figure}[t!]
  \centering
  \includegraphics[width=0.8\columnwidth]{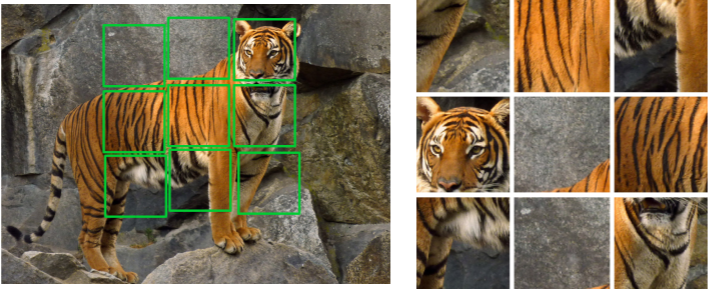}
  \vspace{-0.1cm}
  \caption{Example of the jigsaw puzzle task for a given unlabeled image\,\cite{DBLP:conf/eccv/NorooziF16}.}
  \label{fig:jigsawnet}
  \vspace{-0.3cm}
\end{figure}

\chgd{We now proceed to unsupervised techniques.} Since there are no labels, we need to develop new tasks exploiting labels that can be obtained from the data for free. JigsawNet\,\cite{DBLP:conf/eccv/NorooziF16} is one of such techniques. If an image is divided into smaller regions without labels, we can randomize the regions and solve the jigsaw puzzle where we know the correct order and positions, as illustrated in Figure~\ref{fig:jigsawnet}. JigsawNet trains a context-free network\,(CFN) to solve the jigsaw task. The trained network can be transferred or fine-tuned to solve the given task using a small amount of labeled data. 

\section{Fair Model Training}
\label{sec:fairtraining}

\revision{We now focus on the issue of model fairness where biased data may cause a model to be discriminating and thus unfair. This problem is closely related to robust model training where instead of addressing noise in the training data, the goal is to address bias.} 
\chgd{A famous example is the COMPAS tool by Northpointe, which predicts a defendant's risk of committing another crime. According to an analysis by ProPublica\,\cite{machinebias}, black defendants are far more likely to be judged as high risk compared to white defendants, which turns out to be inaccurate in practice. Other popular examples include an AI-based recruiting system discriminating against job applicants by gender\,\cite{amazonaibias}, an AI-based photo app tagging people of a certain race inappropriately\,\cite{googleaibias}, and an AI chatbot generating hate speech towards minorities\,\cite{leeludaaibias}. These incidences fueled the new research area of algorithmic fairness. }

There could be multiple reasons why COMPAS discriminates. The training data could be biased where there is more data for certain demographics. Or there can be external factors where the surrounding environment may have caused more crime than race itself. Even the fairness measure can be in question where it does not accurately reflect reality. In general, analyzing fairness can be an extremely complicated issue that involves factors outside the data. 

\chgd{An extensive discussion on fairness and ethics can be found in the recent fair ML book\,\cite{barocas-hardt-narayanan}, and here we only focus on fairness issues with technical solutions. In particular, we discuss how to measure fairness and how to mitigate unfairness. In addition, we discuss a recent trend of how fair and robust techniques are converging. \revision{This trend is natural, as bias and noise can affect each other, and only addressing fairness may negatively affect robustness and vice versa.} 
This section extends recent tutorials\,\cite{DBLP:journals/pvldb/Whang020,DBLP:conf/kdd/0001RSW21} by the authors.}

\subsection{Fairness Measures}
\label{sec:measuringfairness}

Fairness cannot be described by one notion, and there are tens of possible definitions summarized in various surveys\,\cite{barocas-hardt-narayanan,DBLP:conf/pods/Venkatasubramanian19,DBLP:journals/cacm/ChouldechovaR20,DBLP:journals/corr/abs-1908-09635} used for predicting crime, hiring, giving loans, and more. \chgd{We illustrate representative measures using a running example and then categorize them according to reference\,\cite{barocas-hardt-narayanan} as shown in Table~\ref{tbl:alltechniques} on Page~\pageref{tbl:alltechniques}.} We use the following notations: $Y$ denotes the label of a sample, $\hat{Y}$ the prediction of a model, and $Z$ is a sensitive attribute like race or gender. \chgd{Choosing a sensitive attribute depends on what is considered sensitive in the application. For example, if a company may run into trouble by discriminating based on age, then an attribute that is related to age can be considered sensitive. }

\begin{figure}[h]
  \includegraphics[width=\columnwidth]{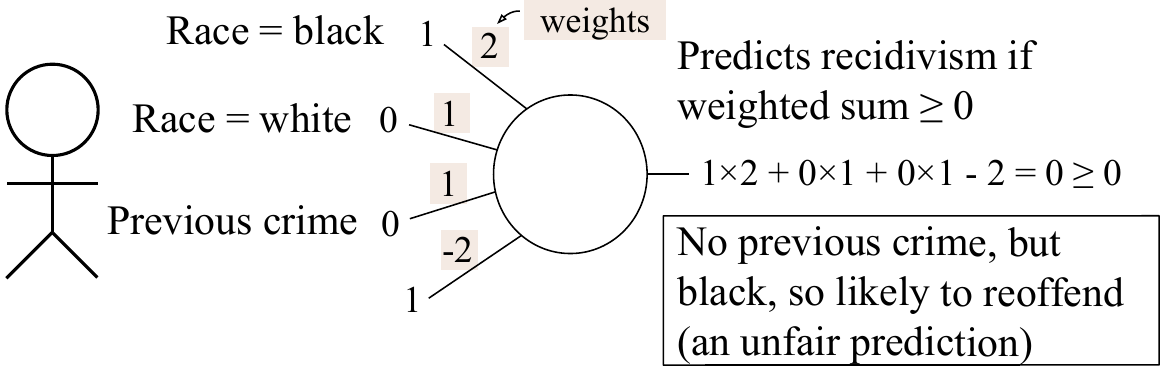}
  \caption{A perceptron receiving input features and performing a weighted sum. Even a model as simple as a perceptron may show unfairness.}
  \label{fig:perceptron}
\end{figure}

We illustrate fairness using the simplest-possible model: a perceptron, which is the most basic unit in a neural network. \chgd{Suppose the perceptron receives three input features: ``Race = black'' has a value of one if the person is black (e.g., $Z$ = 0) or zero otherwise ($Z$ = 1). ``Race = white'' has one if it is a white person or zero otherwise.} 
``Previous crime'' is one if the person has a previous crime and zero otherwise. The last feature is a constant to make the prediction threshold equal to zero. \chgd{Given an example, we take the weighted sum by multiplying the feature values with the weights [2, 1, 1, -2] and, if the sum is at least zero, the model predicts (i.e., the $\hat{Y}$ value) recidivism (i.e., $Y$ = 1) and otherwise not (i.e., $Y$ = 0).} 
For example, if a white person committed a previous crime, the weighted sum is $0 \times 2 + 1 \times 1 + 1 \times 1 - 1 \times 2 = 0$, which is larger or equal to the threshold zero. The interpretation is that the person previously committed a crime, so is likely to re-offend. A black person who committed a previous crime gets the same prediction. However, for people who did not commit a previous crime, the model starts to discriminate where only a black person is predicted to still re-offend as shown in Figure~\ref{fig:perceptron}. The prediction is obviously unfair and is shown for illustration purposes to show how even a model as simple as a perceptron can be discriminating.

For our running example, let us assume there are four people: ($a$) one white person who committed a crime before and committed a crime again, ($b$) one white person who never committed a crime, ($c$) one black person who committed a crime before and committed a crime again, and ($d$) one black person who never committed a crime.
In this example, the perceptron correctly predicts for $a$, $b$, and $c$ but wrongly predicts for $d$.

We summarize the prominent group fairness measures for fairness.

\chgd{
\begin{itemize}
    \item \emph{Demographic parity}\,\cite{DBLP:conf/kdd/FeldmanFMSV15,DBLP:conf/innovations/DworkHPRZ12} requires that sensitive groups must have the same positive prediction rates. The formulation is as follows: $P(\hat{Y} = 1 | Z = 0) = P(\hat{Y} = 1 | Z = 1)$ where the $Z$ value indicates the sensitive group. $\hat{Y}$ = 1 means that the prediction of the model is positive, e.g., predicting recidivism. Demographic parity says that the positive prediction rates of the two groups must be the same. In our running example, $P(\hat{Y} = 1 | Z = 0) = 0.5$ while $P(\hat{Y} = 1 | Z = 1) = 1$, which shows unfairness. 
    \item \emph{Equalized odds}\,\cite{DBLP:conf/nips/HardtPNS16,DBLP:conf/www/ZafarVGG17,berk2017fairness} is defined as $P(\hat{Y} = 1 | Z = 0, Y = A) = P(\hat{Y} = 1 | Z = 1, Y = A), A \in \{0, 1\}$. That is, we would like to guarantee demographic parity when the label $Y$ is zero (in our example, the person did not commit crime again) and when $Y$ is one (the person committed crime) separately. In other words, equalized odds says that the accuracy conditioned on the true label must be the same for the groups. In our running example, $P(\hat{Y} = 1 | Z = 0, Y = 1) =  P(\hat{Y} = 1 | Z = 1, Y = 1) = 1$, but $P(\hat{Y} = 1 | Z = 0, Y = 0) = 0 \neq  P(\hat{Y} = 1 | Z = 1, Y = 0) = 1$, so there is some unfairness.
    \item \emph{Predictive parity}\,\cite{DBLP:journals/bigdata/Chouldechova17,2016COMPASRS,berk2017fairness} is defined as $P(Y = 1 | Z = 0, \hat{Y} = 1) = P(Y = 1 | Z = 1, \hat{Y} = 1)$. That is, given that the predictions are positive, we would like the actual likelihood of the label being positive to also be the same. Note that this measure can be extended to other label classes (e.g., $Y = 0, \hat{Y} = 0$). In our running example, $P(Y = 1 | Z = 0, \hat{Y} = 1) = 1 \neq P(Y = 1 | Z = 1, \hat{Y} = 1) = 0.5$, which shows unfairness.
\end{itemize}
}

Interestingly, many statistical fairness measures are equivalent to or variants of the following fairness criteria\,\cite{barocas-hardt-narayanan}: independence: $\hat{Y} \bot Z$, separation: $\hat{Y} \bot Z | Y$, and sufficiency: $Y \bot Z | \hat{Y}$. Note that demographic parity is equivalent to independence, equalized odds is equivalent to separation, and predictive parity is equivalent to sufficiency. An impossibility result says that no two fairness criteria can be fully satisfied together (see proofs in \,\cite{barocas-hardt-narayanan}).

There are remaining fairness criteria beyond the three above, and we cover the two popular ones: individual fairness and causality fairness.

\chgd{
\begin{itemize}
    \item \emph{Individual fairness}\,\cite{DBLP:conf/innovations/DworkHPRZ12} only uses the classifier for its definition and is defined as $D(f(x), f(x^\prime)) \leq d(x, x^\prime)$ where $d$ is a distance function among examples, and $D$ is a distance function between outcome distributions. Intuitively, the predictions for similar people must be similar as well. For example, if two people are similar to each other, then their recidivism rates must be similar as well. Choosing proper distance functions is a key challenge in individual fairness.
    \item \emph{Causality fairness}\,\cite{10.5555/3294771.3294834,NIPS2017_6995,Zhang2018FairnessID,Nabi2018FairIO,10.1145/3308558.3313559} assumes a causal model, which is a diagram of relationships between attributes. An edge from attribute $A$ to attribute $B$ means that $A$'s value affects $B$'s value. For example, suppose that race not only affects crime, but also the zip code of a person's address, which provides an environment for committing more or less crime. A causal graph could have three nodes {\sf race}, {\sf zip code}, and {\sf crime} with edges from {\sf race} to {\sf zip code}, {\sf race} to {\sf crime}, and {\sf zip code} to {\sf crime}. One can perform a counterfactual analysis to see if zip code indeed affects crime rates by comparing similar people that live or do not live in that zip code. 
\end{itemize}
}

\vspace{-0.2cm}
\subsection{Unfairness Mitigation}
\label{sec:unfairnessmitigation}

Although there are many ways to measure fairness, one would ultimately like to perform {\em unfairness mitigation}\,\cite{barocas-hardt-narayanan,DBLP:journals/ibmrd/BellamyDHHHKLMM19}. Data bias can be addressed either before, during, or after model training. \chgd{These approaches are referred to as {\em pre-processing}, {\em in-processing}, and {\em post-processing} approaches respectively.} Pre-processing approaches can be viewed as data cleaning, but with a focus on improving fairness. For each approach, we cover representative techniques.

\paragraph{Pre-processing Mitigation}

The goal is to fix the unfairness before model training by removing data bias. The advantage is that we may be able to solve the root cause of unfairness within the data. A disadvantage is that it may be tricky to ensure that the model fairness actually improves when we only operate on the data. A na\"ive approach that does not work is to remove sensitive attributes (also referred to as unawareness) because they are usually correlated with other attributes. For example, removing sensitive attributes like race, income, and gender does not ensure fairness because their values can be inferred using correlated attributes like zip code, credit score, and browsing history, respectively. \chgd{We cover three natural approaches for pre-processing -- repairing data, generating data, and acquiring data.}

\begin{figure}[h]
\centering
  \includegraphics[width=0.8\columnwidth]{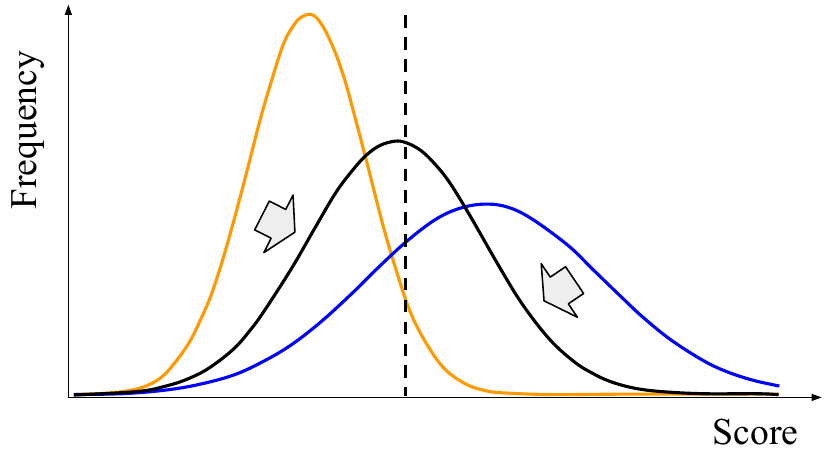}
  \caption{Repairing distribution by averaging scores of the same percentiles\,\cite{DBLP:conf/kdd/FeldmanFMSV15}.}
  \label{fig:datarepair}
  \vspace{-0.2cm}
\end{figure}

\chgd{For fair data repairing, we first cover a method\,\cite{DBLP:conf/kdd/FeldmanFMSV15} that guarantees demographic parity while preserving important statistics like the ranking of data. As an example (Figure~\ref{fig:datarepair}), let us say that we have test scores and distributions for different genders where the data follow normal distributions, but the women's distribution has a higher mean and larger variance.} Now let us say that we want to train a model that uses the test score to make a prediction. If we keep the men and women distributions as they are, then a model using a single threshold is going to be unfair for male versus female and violate demographic parity. Hence, the idea is to combine the two distributions by averaging the scores of the same percentile without losing the ranking information. This method can be extended to more than two sensitive groups. 

A more recent system called Cappucin\,\cite{DBLP:conf/sigmod/SalimiRHS19} repairs data such that a new causality-based fairness called interventional fairness is satisfied. The key insight is that satisfying interventional fairness can be reduced to satisfying multivalued functional dependencies (MVDs). The authors then propose minimal repair methods for MVDs by reducing the problem to MaxSAT or matrix factorization problems.

If there is not enough data to satisfy fairness, an alternative is to generate new data using the available data. A recent method\,\cite{DBLP:conf/icml/ChoiGSSE20} is to generate unbiased data using weak supervision. The input is biased data and an unbiased data that is smaller than the biased data that we have some control on. The idea is to train a generative model on the bias data except that we are adjusting the example weights such that it is as if the generative model is being trained on unbiased data. Then the generative model generates new data that is unbiased. An example weight reflects how likely the example is part of the biased or unbiased data and can be computed by training a separate classifier for distinguishing the biased data from the unbiased data. The generative model that uses the example weights for training is guaranteed to produce unbiased synthetic data. In addition, GANs\,\cite{DBLP:conf/bigdataconf/XuYZW18} have also been used for data generation where a generator competes with two discriminators: one for telling apart real and fake data and another for predicting the sensitive attribute.

\begin{figure}[h]
  \includegraphics[width=0.95\columnwidth]{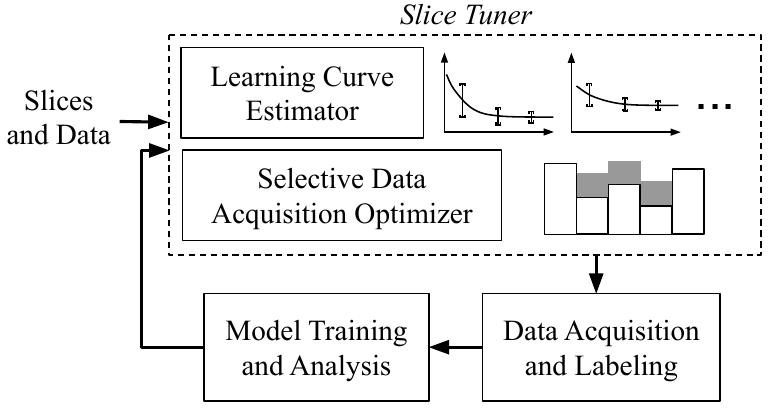}
  \caption{Slice Tuner\,\cite{DBLP:conf/sigmod/TaeW21} is a selective acquisition framework for accurate and fair models where it iteratively estimates learning curves to determine how much data to acquire per data slice.}
  \label{fig:slicetuner}
  \vspace{-0.3cm}
\end{figure}

As data is increasingly available, acquiring data from external data sources is also becoming a viable option\,\cite{DBLP:conf/nips/ChenJS18,DBLP:conf/icde/AsudehJJ19}. A recent approach called Slice Tuner\,\cite{DBLP:conf/sigmod/TaeW21} selectively acquires examples with the purpose of maximizing both accuracy and fairness of the trained model (Figure~\ref{fig:slicetuner}). Slice Tuner assumes a set of non-overlapping data slices (e.g., regions), and the fairness measure is equal error rates\,\cite{DBLP:conf/pods/Venkatasubramanian19} where the model's accuracies on different slices must be similar. The key idea is to maintain learning curves of slices, which can be used to predict accuracies on those slices given more data. Slice Tuner then solves a convex optimization problem to determine the amount of data to acquire per slice. Two challenges are that learning curves may be unreliable and that acquiring data for one slice may influence the model's accuracy on another slice. Slice Tuner solves these problems by iteratively updating the learning curves using a proxy for estimating influence. As a result, a model can obtain better accuracy and fairness compared to various baselines given a fixed budget for data acquisition. 
Another system called Deepdiver\,\cite{DBLP:conf/icde/AsudehJJ19} performs data acquisition such that all possible slices contain sufficient amounts of data. Here the slices may overlap with each other, and the objective is to guarantee minimum coverage instead of improving model accuracy or fairness.

\paragraph{In-processing Mitigation}

We now cover representative in-processing techniques for unfairness mitigation where the model training is fixed. The advantage is that one can directly optimize accuracy and fairness. On the other hand, the downside is that the model training itself may have to change significantly, which may not be feasible in many applications. \chgd{There are largely three in-processing approaches.} The first is to directly modify the objective function of the model training by adding fairness constraints. The second is to make the model compete with a fairness discriminator via adversarial training. The third is adaptive sample reweighting techniques that re-weight input samples for fairness.

\begin{figure}[h]
  \centering
  \includegraphics[width=0.7\columnwidth]{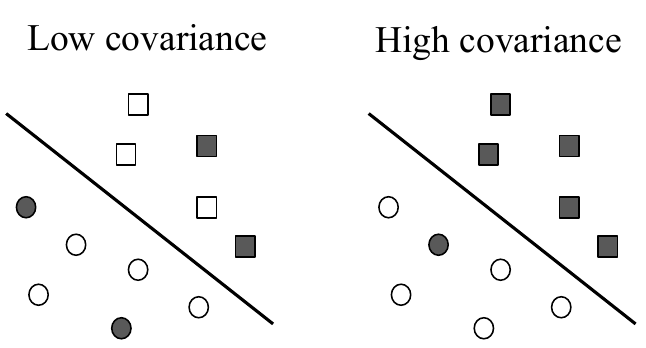}
  \caption{Suppose we are classifying circles versus squares where the color of the shapes indicate the sensitive group. The left image has little correlation between the sensitive group and being on one side of the decision boundary, which means the covariance is low. The right image, on the other hand, shows a high covariance where just by looking at the sensitive group, one can figure out whether the data point is on which side of the decision boundary. Hence, by minimizing the covariance, one can also make the sensitive attribute more independent of the model predictions and thus satisfy demographic parity as well.}
  \label{fig:fairnessconstraints}
  \vspace{-0.3cm}
\end{figure}

\chgd{Directly adding fairness constraints to the model training objective function is an effective way to optimize for fairness.} Zafar et al. propose fairness constraints techniques\,\cite{DBLP:conf/aistats/ZafarVGG17} to use in the objective function of model training to satisfy demographic parity. The focus is on convex margin classifiers like SVMs. However, as the demographic parity constraint is not convex, it cannot be directly added to the objective function. Instead, the idea is to use a proxy that approximates demographic parity and is convex. For the proxy, the authors use the covariance between the sensitive attribute and the signed distance to the decision boundary. Figure~\ref{fig:fairnessconstraints} provides an intuition why covariance is a good proxy. A limitation of fairness constraints is that it does not readily generalize to deep neural networks that are not convex. Other optimization techniques\,\cite{DBLP:conf/icml/AgarwalBD0W18,DBLP:conf/pkdd/KamishimaAAS12} for maximizing fairness and accuracy have been proposed as well.

\begin{figure}[h]
  \centering
  \includegraphics[width=\columnwidth]{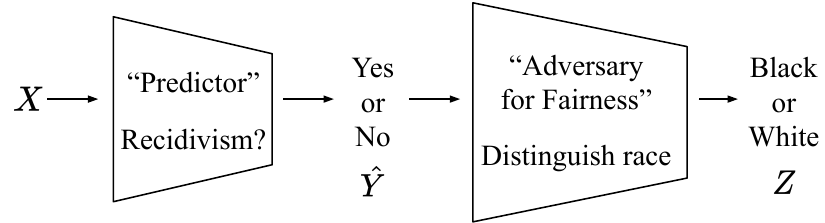}
  \caption{Adversarial de-biasing\,\cite{DBLP:conf/aies/ZhangLM18} competes a classifier with a fair discriminator.}
  \label{fig:adversarialdebiasing}
  \vspace{-0.3cm}
\end{figure}

If one does not want to modify the loss function in the model, another approach is to perform adversarial training with another model for fairness.
Adversarial de-biasing\,\cite{DBLP:conf/aies/ZhangLM18} is a representative work in this direction. Here the idea is to do adversarial training between a binary classifier and an adversary that tries to infer the sensitive attribute value (Figure~\ref{fig:adversarialdebiasing}). For example, the classifier may predict recidivism while the adversary infers the gender of the person based on the classifier predictions. Suppose the fairness measure is demographic parity. A key theoretical result is that, if the adversary optimally predicts the sensitive attribute, but the classifier completely fools the adversary, it means that the model prediction is independent of the sensitive attribute. In our example, recidivism will have nothing to do with the gender. One downside of adversarial debiasing or adversarial training in general is that stability is sometimes an issue where the model training may not easily converge to a single solution.

Adversarial training can also be used to attain both fair and robust training. FR-Train\,\cite{DBLP:journals/corr/abs-2002-10234} uses a mutual information-based approach to train a model that is both fair and robust. The classifier's fairness-accuracy tradeoff is harmed when the data is poisoned. FR-train avoids this problem by competing a classifier with two discriminators for fairness and robustness. The robustness discriminator uses a clean validation set that can be constructed using crowdsourcing techniques. We will continue discussing this work in Section~\ref{sec:robustandfair}.

\begin{figure}[h]
\centering
  \includegraphics[width=0.95\columnwidth]{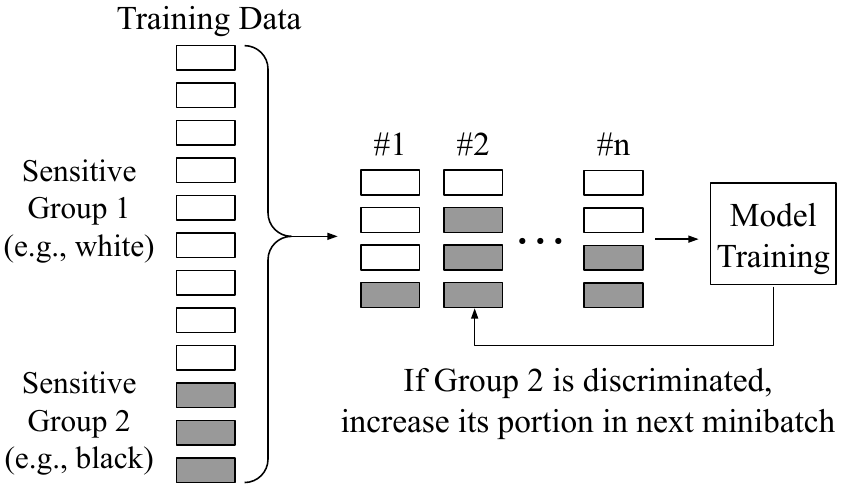}
  \caption{FairBatch\,\cite{DBLP:conf/iclr/Roh0WS21} is a batch selection framework for model fairness where sensitive group ratios are adjusted based on intermediate model fairness.}
  \label{fig:fairbatch}
  \vspace{-0.4cm}
\end{figure}

\chgd{A major downside of the previous methods is that the model training needs to be replaced or modified significantly, and a more convenient approach is to only re-weight the samples in order to obtain similar fairness results.} FairBatch\,\cite{DBLP:conf/iclr/Roh0WS21} is a batch selection technique with the purpose of improving fairness. During batch selection, it is common to select a random sample from the training set. Instead, the idea of FairBatch is to adjust the sensitive group ratios within each batch of examples being used for training as illustrated in Figure~\ref{fig:fairbatch}. For example, suppose the training set is biased where a certain sensitive group has very few examples. If an intermediate model shows poor fairness, then the next batch of examples will contain more examples of that sensitive group.
How exactly the sensitive group ratio should be adjusted is the technical challenge. 
OmniFair\,\cite{DBLP:conf/sigmod/ZhangCAN21} is a declarative system for supporting group fairness for any model by reweighting samples. While the goals are similar to FairBatch, the specific optimization techniques differ where Omnifair uses a Lagrangian multiplier to translate a constraint optimization problem into an unconstrained optimization problem and leverages a monotonicity property. Other techniques include an adaptive sample reweighting approach that corrects label bias\,\cite{DBLP:conf/aistats/JiangN20} and an adaptive boosting technique for maximizing fairness\,\cite{DBLP:conf/cikm/IosifidisN19}.

\paragraph{Post-processing Mitigation}

\chgd{The final approach for unfairness mitigation is to fix model predictions for fairness, which is the only option if the data and model cannot be modified. However, post-processing usually results in a tradeoff of worse accuracy. }

\begin{figure}[h]
  \includegraphics[width=\columnwidth]{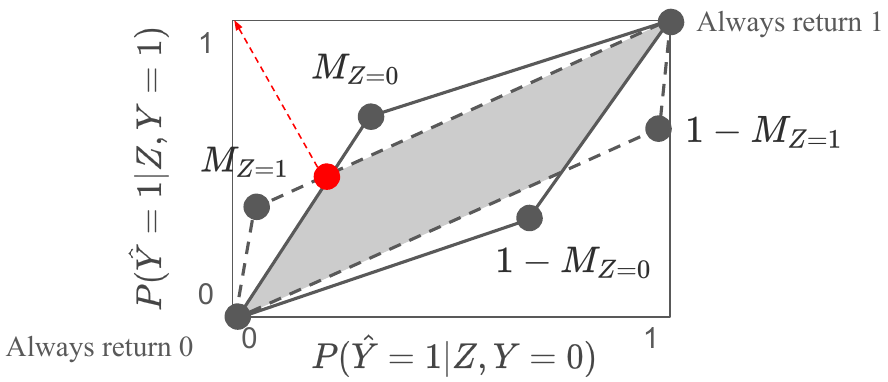}
  \caption{Post-processing unfairness mitigation\,\cite{DBLP:conf/nips/HardtPNS16} involves combining models using randomization to attain the desired fairness.}
  \label{fig:hardt}
\end{figure}

We introduce a representative work\,\cite{DBLP:conf/nips/HardtPNS16} that combines models to adjust fairness. The method we explain here assumes equalized odds for binary classifiers, although other settings are supported in the paper as well. We assume a model $M$ for each $Z$ value and then construct the following models: a trivial model that only returns 0, another trivial model that only returns 1, and the ``inverted'' model $1 - M$, which returns the opposite prediction of $M$. The idea is to combine these models using randomization such that the fairness criteria is satisfied. Figure~\ref{fig:hardt} illustrates how this combination can be done for $Z = 0$ and $Z = 1$. In each case, we can generate a model with the desired positive prediction rate as long as it is inside the parallelogram. If we generate a model in the intersection of the two parallelograms, we can find a model where $P(\hat{Y} = 1 | Z = 0, Y = A) = P(\hat{Y} = 1 | Z = 1, Y = A), A \in \{0, 1\}$, which is exactly the definition of equalized odds. Among the possible combined models, we then choose the one with the lowest expected loss (i.e., closest to the top left as highlighted in the figure). Other post-processing approaches leverage unlabeled data\,\cite{DBLP:conf/nips/ChzhenDHOP19} and calibration\,\cite{DBLP:conf/nips/PleissRWKW17}.

\subsection{Convergence with Robustness Techniques}
\label{sec:robustandfair}

\begin{figure}[h]
  \centering
  \includegraphics[width=0.86\columnwidth]{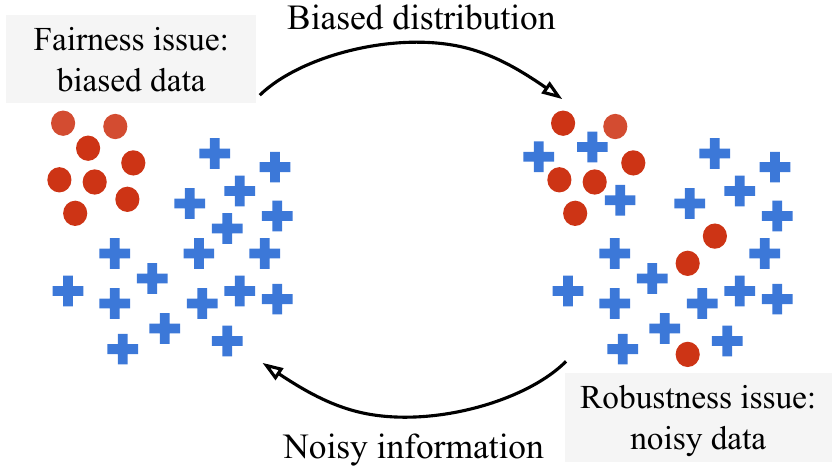}
  \caption{Fairness and robustness issues may negatively affect each other. Noisy or missing group information may result in inaccurate results after unfairness mitigation. A biased distribution in the data may result in disproportionate accuracies after robust training.}
  \label{fig:fairandrobust}
  \vspace{-0.4cm}
\end{figure}

Most recently, we are witnessing a convergence of fairness and robustness techniques. This direction is inevitable because both techniques address flaws in the data, but one does not subsume the other. Fair training assumes that the data is clean and only focuses on removing its bias. However, the sensitive attribute itself can be noisy or even missing. \chgd{On the other hand, robust training primarily focuses on improving the overall accuracy, but does not consider disproportionate performances between different sensitive groups. In general, fairness and robustness are not necessarily aligning goals. For example, if the data is already biased, then removing noisy data for robust training may end up worsening the bias by removing too much data from an underrepresented group\,\cite{roh2021sample}.} Figure~\ref{fig:fairandrobust} illustrates these dynamics. There are three directions for the convergence: making fairness approaches more robust (fairness-oriented), making robust approaches fairer (robust-oriented), and equal mergers of fair and robust training. We summarize the recent research for each of the three approaches.

\paragraph{Fairness-oriented Approaches}

The first direction of convergence is to make fair training more robust. This research currently has two directions: when the sensitive group information is noisy or entirely missing. 

The first scenario may occur if some users may want to hide or mistakenly omit their group memberships. An analysis of fair training results on noisy sensitive group information\,\cite{DBLP:conf/nips/WangGNCGJ20} shows that the true fairness violation on a clean sensitive group can be bounded by a distance between this group and its noisy version. In addition, noise-tolerant fair training techniques\,\cite{DBLP:conf/nips/LamyZ19} have been proposed where the idea is to change the unfairness tolerance to estimate the fairness of the true data distribution. 

\begin{figure}[h]
  \includegraphics[width=0.9\columnwidth]{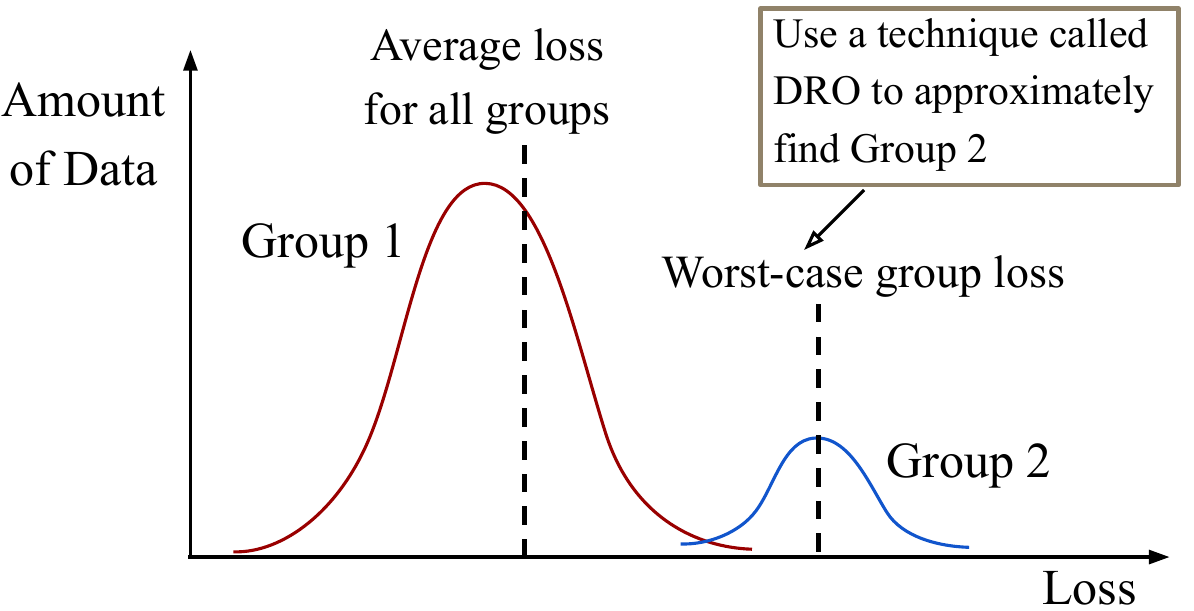}
  \caption{A DRO-based fair algorithm\,\cite{DBLP:conf/icml/HashimotoSNL18} improves model fairness without using the sensitive group information by identifying the worst-performing samples.}
  \label{fig:dro}
  \vspace{-0.2cm}
\end{figure}

The second scenario is when the sensitive attribute is fully missing. Here the data collection sometimes does not gather the group information due to various reasons like legal restrictions.
Distributionally Robust Optimization (DRO)\,\cite{Sinha2017CertifyingSD} has been used to improve the model performance for minority sensitive groups without using the group information\,\cite{DBLP:conf/icml/HashimotoSNL18}. 
The idea is to approximately minimize the worst-case (latent) group loss by identifying the worst-performing samples (Figure~\ref{fig:dro}) and giving them more weight. Adversarially-reweighted learning for fairness\,\cite{DBLP:conf/nips/LahotiBCLPT0C20} makes the assumption that unobserved sensitive attributes are correlated with the features and labels, and performs adversarial training between a classifier versus an adversary that finds less accurate clustered regions and gives more weights on those regions.

\paragraph{Robustness-oriented Approaches}

\begin{figure}[h]
\vspace{-0.3cm}
\centering
  \includegraphics[width=\columnwidth]{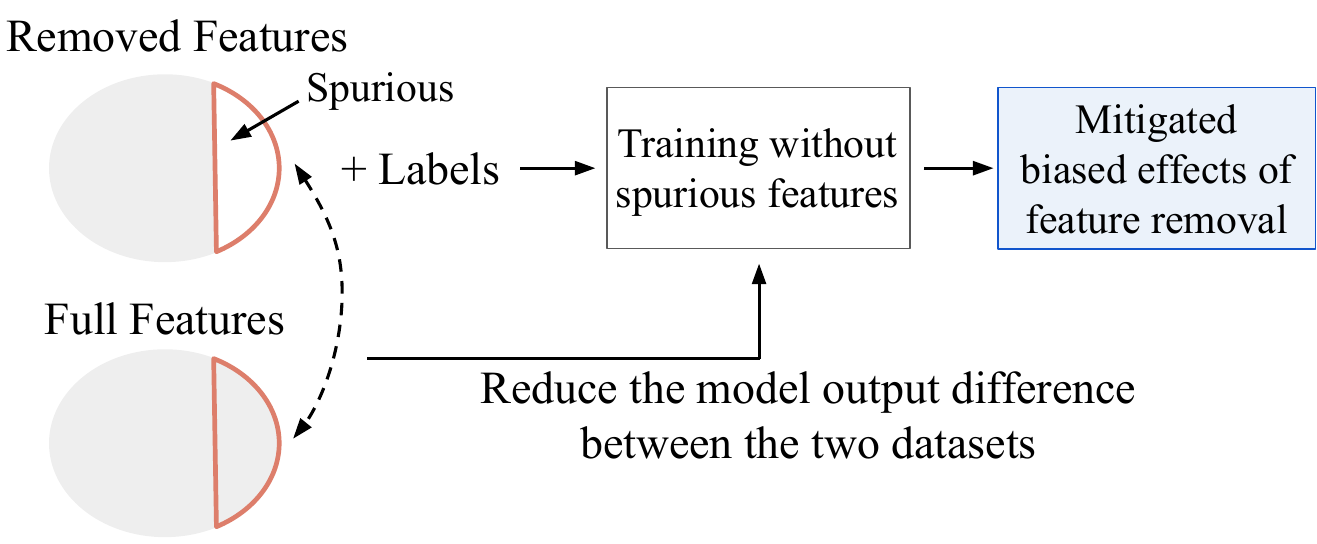}
  \caption{A self-training technique\,\cite{DBLP:conf/fat/KhaniL21} can mitigate the biased effects of spurious feature removal by also using full-featured data.}
  \label{fig:fairremoval}
  \vspace{-0.5cm}
\end{figure}

Robust training is designed to improve the overall accuracy of a model, but may discriminate groups where some have much worse accuracy than others. There are three directions of research: finding anomalies in the data, training without spurious features, and improving robustness via adversarial training. Fair anomaly detection\,\cite{DBLP:conf/fat/ZhangD21} has been proposed to prevent anomaly detection from discriminating specific groups. The idea is to compete a classifier that finds abnormal data and a discriminator that predicts the sensitive group from the classifier's prediction. After training, the classifier's output becomes independent of the sensitive group. Fair training without spurious features\,\cite{DBLP:conf/fat/KhaniL21} addresses the problem of preventing feature removal from being discriminating. A self-training technique is proposed to mitigate accuracy degradation and biased effects (Figure~\ref{fig:fairremoval}). Finally, fair adversarial training\,\cite{DBLP:conf/icml/XuLLJT21} prevents adversarial training from discriminating groups by adding constraints for equalizing accuracy and robustness.

\paragraph{Equal Mergers}

\begin{figure}[h]
  \hspace{-0.15cm}
  \includegraphics[width=1.0\columnwidth]{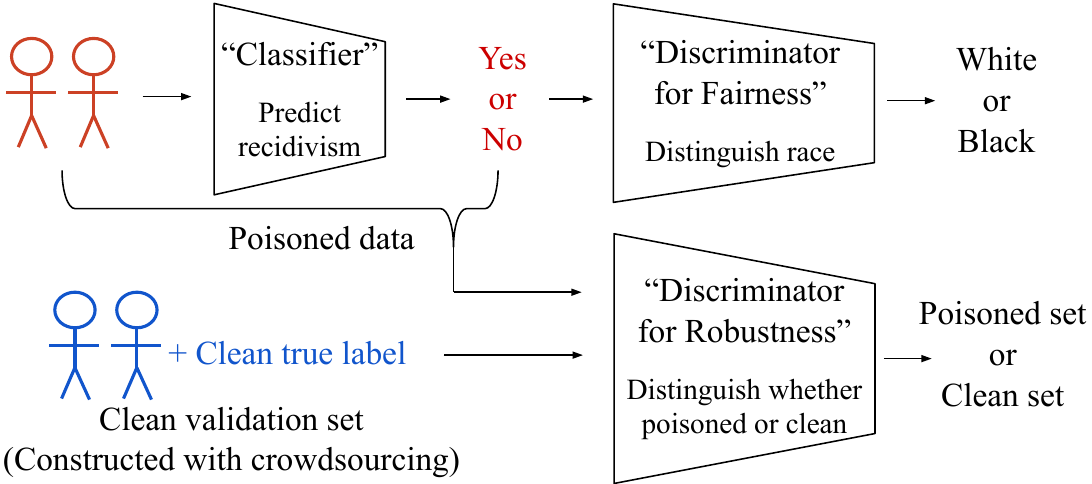}
  \caption{FR-Train\,\cite{DBLP:journals/corr/abs-2002-10234} is a mutual information-based approach for achieving both fairness and robustness, which competes one classifier with two discriminators.}
  \label{fig:frtrain}
   \vspace{-0.3cm}
\end{figure}

\begin{figure}[t]
\hspace{-0.3cm}
  \includegraphics[width=1.05\columnwidth]{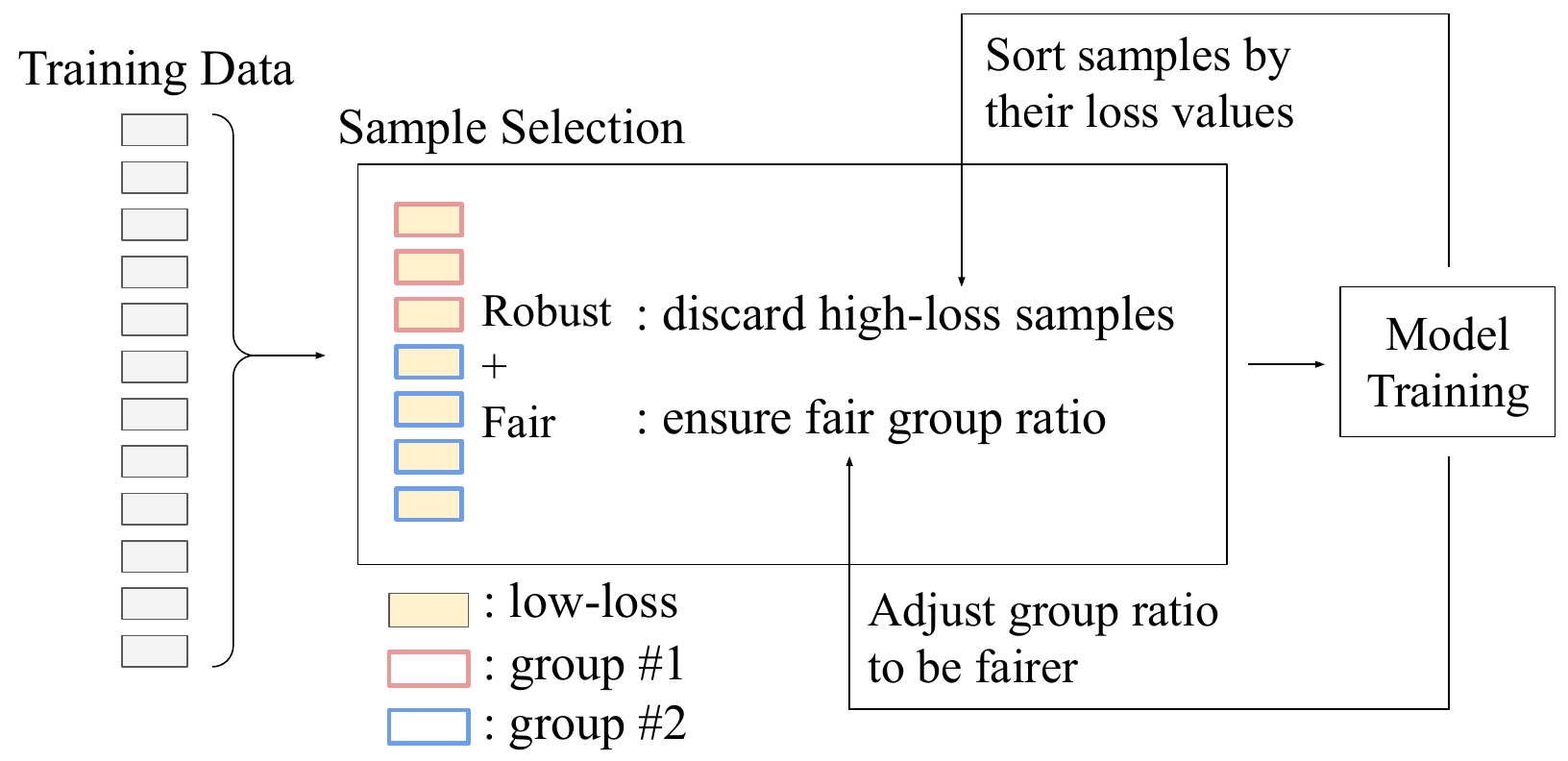}
  \caption{Adaptive sample selection\,\cite{roh2021sample} can be another solution for improving fairness and robustness. The key idea is to utilize only clean and fair samples in training.}
  \label{fig:fairrobustselection}
  \vspace{-0.2cm}
\end{figure}

Robust and fair training can be combined in equal terms as well. One direction is to make the model training fair and robust at the same time. FR-Train\,\cite{DBLP:journals/corr/abs-2002-10234} is a mutual information-based framework that competes a classifier, discriminator for fairness, and discriminator for robustness to make the classifier fair and robust (Figure~\ref{fig:frtrain}). A recent sample selection framework\,\cite{roh2021sample} adaptively selects training samples for fair and robust model training (Figure~\ref{fig:fairrobustselection}). This framework does not require modifying the model or leveraging additional clean data. A fairness-aware ERM framework\,\cite{DBLP:conf/fat/WangLL21} has been proposed based on the observation that group-dependent label noises may reduce both model accuracy and fairness. The solution is to use surrogate loss where the label distribution is corrected based on the noise rates of groups. The surrogate loss better reflects the true loss and thus mitigates the negative effects of group-dependent label noises. Another direction of robust and fair training is to take a role of an adversary and generate attacks that not only reduce accuracy, but also harm fairness. Fairness-targeted poisoning attacks\,\cite{DBLP:conf/pkdd/SolansB020} proposes a gradient-based attack method that finds the optimal attack locations that reduce the fairness the most.

\vspace{-0.3cm}
\chgd{
\section{Overall Findings and Future Directions}
\label{sec:overall_findings}

We summarize our findings. In Section~\ref{sec:datacollection}, we explained that data collection techniques consist of data acquisition, data labeling, and improving existing data and models. Some of the techniques have been studied by the data management community while others by the machine learning community. In Section~\ref{sec:datavalidationandcleaning}, we covered key approaches in data validation, data cleaning, data sanitization, and data integration. Data validation can be performed using visualizations and schema information. Data cleaning has been heavily studied where recent techniques are more tailored to improving model accuracy. Data sanitization has the different flavor of defending against poisoning attacks. Data integration is challenging due to multimodal data. In Section~\ref{sec:robusttraining}, we explained that noisy or missing labels incur poor generalization on test data. Existing work for noisy labels suffers from either (i) accumulated noise or (ii) partial exploration of training data. Hybrid\,(e.g., SELFIE) and semi-supervised techniques\,(e.g., DivideMix) can achieve very high accuracy with noisy training data. Semi-supervised\,(e.g., MixMatch) and self-supervised\,(e.g., JigsawNet) techniques are actively developed to exploit abundant unlabeled data. In Section~\ref{sec:fairtraining}, we covered fairness measures, unfairness mitigation techniques, and convergence with robustness techniques. The mitigation can be done before, during, or after model training. Pre-processing is useful when training data can be modified. In-processing is useful when the training algorithm can be modified. Post-processing can be used when we cannot modify the data and model training. The convergence with robustness techniques can be categorized into fair-robust techniques, robust-fair techniques, and equal mergers.

As data-centric AI becomes more established we believe there will be various convergences of these research areas. Our list is certainly not exhaustive, but we attempt to identify the major trends.

\begin{itemize}
    \item {\em Data cleaning and robust training}: Currently data cleaning is becoming more machine learning oriented, but is considered less effective than robust training. We believe that the two techniques should continue integrating for the best results.
    \item {\em Data validation and model fairness}: The recent works in data validation point to AI ethics as one of the challenging aspects to validate. We believe that model fairness will eventually be merged into the data validation process.
    \item {\em \revision{Data collection}}: So far, most of the machine learning literature assumes that the input data is already given. At the same time, data collection for accurate machine learning is now an active research direction in the data management community. We believe this trend will continue to expand where data collection needs to also consider fairness and robustness.
    \item {\em Model training and testing}: Improving model training and testing protocols is becoming another solution for dealing with data quality issues. The output of the model on data samples provides useful knowledge for evaluating the data, helping to develop accurate and robust inference pipelines. We believe that the learning dynamics of models provide new perspectives for interpreting robustness and fairness.
    \item {\em Model fairness and robustness}: Trustworthy AI is becoming increasingly critical in the machine learning community, and we believe its various aspects including fairness and robustness will have to be addressed together instead of one at a time. There are other elements of Trustworthy AI including privacy and explainability that should eventually be part of data-centric AI as well.
\end{itemize}}

\paragraph{\revision{Concluding Remark}}
In the data-centric AI era, collecting data and improving its quality will only become more critical for deep learning. We covered four major topics (data collection, data cleaning, validation, and integration, robust model training, and fair model training), which have been studied by different communities, but need to be used together. We believe all the data techniques will eventually converge with the robust and fair training techniques as data-centric AI matures, and hope that our survey plays a catalyst role.




\begin{acknowledgements}
This work was supported by the Institute of Information \& Communications Technology Planning \& Evaluation(IITP) grant funded by the Korea government(MSIT) (No. 2022-0-00157 and 2020-0-00862) and the National Research Foundation of Korea(NRF) grant funded by the Korea government(MSIT) (No. NRF-2018R1A5A1059921 and NRF-2022R1A2C2004382).
\end{acknowledgements}

\balance

\bibliographystyle{plain}      
\bibliography{main}

\begin{thebibliography}{100}

\bibitem{amazonmechanicalturk}
{Amazon Mechanical Turk}.
\newblock \url{https://www.mturk.com/}.
\newblock Accessed July 13th, 2022.

\bibitem{amazonsagemaker}
{Amazon SageMaker Ground Truth}.
\newblock \url{https://aws.amazon.com/sagemaker/groundtruth/}.
\newblock Accessed July 13th, 2022.

\bibitem{amazonaibias}
{Amazon scraps secret AI recruiting tool that showed bias against women}.
\newblock
  \url{https://www.reuters.com/article/us-amazon-com-jobs-automation-insight-idUSKCN1MK08G}.
\newblock Accessed July 13th, 2022.

\bibitem{crowdflower}
{CrowdFlower Data Science Report}.
\newblock
  \url{https://visit.figure-eight.com/rs/416-ZBE-142/images/CrowdFlower_DataScienceReport_2016.pdf}.

\bibitem{idc}
Data age 2025.
\newblock \url{https://www.seagate.com/our-story/data-age-2025/}.

\bibitem{datacentricai}
Data-centric {AI} resource hub.
\newblock \url{https://datacentricai.org/}.

\bibitem{datanami}
Data prep still dominates data scientists’ time, survey finds.
\newblock
  \url{https://www.datanami.com/2020/07/06/data-prep-still-dominates-data-scientists-time-survey-finds/}.

\bibitem{facets}
Facets -- visualization for {ML} datasets.
\newblock \url{https://pair-code.github.io/facets/}.
\newblock Accessed July 13th, 2022.

\bibitem{gcp}
{GCP} {AI} platform data labeling service.
\newblock \url{https://cloud.google.com/ai-platform/data-labeling/docs}.
\newblock Accessed July 13th, 2022.

\bibitem{googleaibias}
{Google apologises for Photos app's racist blunder}.
\newblock \url{https://www.bbc.com/news/technology-33347866}.
\newblock Accessed July 13th, 2022.

\bibitem{kaggle}
Kaggle.
\newblock \url{https://www.kaggle.com}.

\bibitem{samsung}
Principles for {AI} ethics.
\newblock \url{https://research.samsung.com/artificial-intelligence}.
\newblock Accessed July 13th, 2022.

\bibitem{google}
Responsible {AI} practices.
\newblock \url{https://ai.google/responsibilities/responsible-ai-practices}.
\newblock Accessed July 13th, 2022.

\bibitem{microsoft}
Responsible {AI} principles from {Microsoft}.
\newblock \url{https://www.microsoft.com/en-us/ai/responsible-ai}.
\newblock Accessed July 13th, 2022.

\bibitem{karpathi}
Software 2.0.
\newblock \url{https://medium.com/@karpathy/software-2-0-a64152b37c35}.

\bibitem{leeludaaibias}
{South Korean AI chatbot pulled from Facebook after hate speech towards
  minorities}.
\newblock
  \url{https://www.theguardian.com/world/2021/jan/14/time-to-properly-socialise-hate-speech-ai-chatbot-pulled-from-facebook}.
\newblock Accessed July 13th, 2022.

\bibitem{ibm}
Trusting {AI}.
\newblock
  \url{https://www.research.ibm.com/artificial-intelligence/trusted-ai/}.
\newblock Accessed July 13th, 2022.

\bibitem{DBLP:conf/icml/AgarwalBD0W18}
Alekh Agarwal, Alina Beygelzimer, Miroslav Dud{\'{\i}}k, John Langford, and
  Hanna~M. Wallach.
\newblock A reductions approach to fair classification.
\newblock In {\em ICML}, pages 60--69, 2018.

\bibitem{DBLP:conf/sigmod/AgrawalABBGGLMP19}
Pulkit Agrawal, Rajat Arya, Aanchal Bindal, Sandeep Bhatia, Anupriya Gagneja,
  Joseph Godlewski, Yucheng Low, Timothy Muss, Mudit~Manu Paliwal, Sethu Raman,
  Vishrut Shah, Bochao Shen, Laura Sugden, Kaiyu Zhao, and Ming{-}Chuan Wu.
\newblock Data platform for machine learning.
\newblock In {\em SIGMOD}, pages 1803--1816, 2019.

\bibitem{DBLP:conf/icse/AmershiBBDGKNN019}
Saleema Amershi, Andrew Begel, Christian Bird, Robert DeLine, Harald~C. Gall,
  Ece Kamar, Nachiappan Nagappan, Besmira Nushi, and Thomas Zimmermann.
\newblock Software engineering for machine learning: a case study.
\newblock In {\em ICSE}, pages 291--300, 2019.

\bibitem{machinebias}
J.~Angwin, J.~Larson, S.~Mattu, and L.~Kirchner.
\newblock Machine bias: {T}here's software used across the country to predict
  future criminals. {A}nd its biased against blacks., 2016.

\bibitem{anwar2019real}
Saeed Anwar and Nick Barnes.
\newblock Real image denoising with feature attention.
\newblock In {\em CVPR}, pages 3155--3164, 2019.

\bibitem{DBLP:conf/icde/AsudehJJ19}
Abolfazl Asudeh, Zhongjun Jin, and H.~V. Jagadish.
\newblock Assessing and remedying coverage for a given dataset.
\newblock In {\em ICDE}, pages 554--565, 2019.

\bibitem{DBLP:conf/sigmod/BachRLLSXSRHAKR19}
Stephen~H. Bach, Daniel Rodriguez, Yintao Liu, Chong Luo, Haidong Shao,
  Cassandra Xia, Souvik Sen, Alexander Ratner, Braden Hancock, Houman Alborzi,
  Rahul Kuchhal, Christopher R{\'{e}}, and Rob Malkin.
\newblock Snorkel drybell: {A} case study in deploying weak supervision at
  industrial scale.
\newblock In {\em SIGMOD}, pages 362--375, 2019.

\bibitem{DBLP:journals/pami/BaltrusaitisAM19}
Tadas Baltrusaitis, Chaitanya Ahuja, and Louis{-}Philippe Morency.
\newblock Multimodal machine learning: {A} survey and taxonomy.
\newblock {\em {IEEE} Trans. Pattern Anal. Mach. Intell.}, 41(2):423--443,
  2019.

\bibitem{barocas-hardt-narayanan}
Solon Barocas, Moritz Hardt, and Arvind Narayanan.
\newblock {\em Fairness and Machine Learning}.
\newblock fairmlbook.org, 2019.
\newblock \url{http://www.fairmlbook.org}.

\bibitem{DBLP:conf/kdd/BaylorBCFFHHIJK17}
Denis Baylor, Eric Breck, Heng{-}Tze Cheng, Noah Fiedel, Chuan~Yu Foo, Zakaria
  Haque, Salem Haykal, Mustafa Ispir, Vihan Jain, Levent Koc, Chiu~Yuen Koo,
  Lukasz Lew, Clemens Mewald, Akshay~Naresh Modi, Neoklis Polyzotis, Sukriti
  Ramesh, Sudip Roy, Steven~Euijong Whang, Martin Wicke, Jarek Wilkiewicz, Xin
  Zhang, and Martin Zinkevich.
\newblock {TFX:} {A} tensorflow-based production-scale machine learning
  platform.
\newblock In {\em KDD}, pages 1387--1395, 2017.

\bibitem{DBLP:journals/ibmrd/BellamyDHHHKLMM19}
Rachel K.~E. Bellamy, Kuntal Dey, Michael Hind, et~al.
\newblock {AI} fairness 360: An extensible toolkit for detecting and mitigating
  algorithmic bias.
\newblock {\em {IBM} Journal of Research and Development}, 2019.

\bibitem{berk2017fairness}
Richard Berk, Hoda Heidari, Shahin Jabbari, Michael Kearns, and Aaron Roth.
\newblock Fairness in criminal justice risk assessments: The state of the art,
  2017.

\bibitem{DBLP:conf/iclr/BerthelotCCKSZR20}
David Berthelot, Nicholas Carlini, Ekin~D. Cubuk, Alex Kurakin, Kihyuk Sohn,
  Han Zhang, and Colin Raffel.
\newblock Remixmatch: Semi-supervised learning with distribution matching and
  augmentation anchoring.
\newblock In {\em ICLR}, 2020.

\bibitem{DBLP:conf/nips/BerthelotCGPOR19}
David Berthelot, Nicholas Carlini, Ian~J. Goodfellow, Nicolas Papernot, Avital
  Oliver, and Colin Raffel.
\newblock Mixmatch: {A} holistic approach to semi-supervised learning.
\newblock In {\em NeurIPS}, pages 5050--5060, 2019.

\bibitem{DBLP:journals/debu/BiessmannGRL021}
Felix Biessmann, Jacek Golebiowski, Tammo Rukat, Dustin Lange, and Philipp
  Schmidt.
\newblock Automated data validation in machine learning systems.
\newblock {\em {IEEE} Data Eng. Bull.}, 44(1):51--65, 2021.

\bibitem{DBLP:conf/pkdd/BiggioCMNSLGR13}
Battista Biggio, Igino Corona, Davide Maiorca, Blaine Nelson, Nedim Srndic,
  Pavel Laskov, Giorgio Giacinto, and Fabio Roli.
\newblock Evasion attacks against machine learning at test time.
\newblock In {\em {ECML} {PKDD}}, pages 387--402. Springer, 2013.

\bibitem{Blum:1998:CLU:279943.279962}
Avrim Blum and Tom Mitchell.
\newblock Combining labeled and unlabeled data with co-training.
\newblock In {\em COLT}, pages 92--100, New York, NY, USA, 1998. ACM.

\bibitem{DBLP:conf/cidr/BoehmADGIKLPR20}
Matthias Boehm, Iulian Antonov, Sebastian Baunsgaard, Mark Dokter, Robert
  Ginth{\"{o}}r, Kevin Innerebner, Florijan Klezin, Stefanie~N. Lindstaedt,
  Arnab Phani, Benjamin Rath, Berthold Reinwald, Shafaq Siddiqui, and
  Sebastian~Benjamin Wrede.
\newblock Systemds: {A} declarative machine learning system for the end-to-end
  data science lifecycle.
\newblock In {\em CIDR}, 2020.

\bibitem{breck2019data}
Eric Breck, Martin Zinkevich, Neoklis Polyzotis, Steven Whang, and Sudip Roy.
\newblock Data validation for machine learning.
\newblock In {\em MLSys}, 2019.

\bibitem{DBLP:conf/www/BrickleyBN19}
Dan Brickley, Matthew Burgess, and Natasha~F. Noy.
\newblock Google dataset search: Building a search engine for datasets in an
  open web ecosystem.
\newblock In {\em WWW}, pages 1365--1375, 2019.

\bibitem{DBLP:journals/pvldb/CafarellaHLMYWW18}
Michael~J. Cafarella, Alon~Y. Halevy, Hongrae Lee, Jayant Madhavan, Cong Yu,
  Daisy~Zhe Wang, and Eugene Wu.
\newblock Ten years of webtables.
\newblock {\em {PVLDB}}, 11(12):2140--2149, 2018.

\bibitem{CambroneroFSM17}
Jos{\'{e}} Cambronero, John~K. Feser, Micah~J. Smith, and Samuel Madden.
\newblock Query optimization for dynamic imputation.
\newblock {\em Proc. {VLDB} Endow.}, 10(11):1310--1321, 2017.

\bibitem{DBLP:journals/corr/abs-1810-00069}
Anirban Chakraborty, Manaar Alam, Vishal Dey, Anupam Chattopadhyay, and Debdeep
  Mukhopadhyay.
\newblock Adversarial attacks and defences: {A} survey.
\newblock {\em CoRR}, abs/1810.00069, 2018.

\bibitem{DBLP:conf/nips/ChangLM17}
Haw{-}Shiuan Chang, Erik~G. Learned{-}Miller, and Andrew McCallum.
\newblock Active bias: Training more accurate neural networks by emphasizing
  high variance samples.
\newblock In {\em NeurIPS}, pages 1002--1012, 2017.

\bibitem{CheEtAl_nature_sr18}
Zhengping Che, Sanjay Purushotham, Kyunghyun Cho, David Sontag, and Yan Liu.
\newblock Recurrent neural networks for multivariate time series with missing
  values.
\newblock {\em Nature Scientific Reports}, 8(1):6085, 2018.

\bibitem{DBLP:conf/sigmod/ChenCDD0HKMMNOP20}
Andrew Chen, Andy Chow, Aaron Davidson, Arjun DCunha, Ali Ghodsi, Sue~Ann Hong,
  Andy Konwinski, Clemens Mewald, Siddharth Murching, Tomas Nykodym, Paul
  Ogilvie, Mani Parkhe, Avesh Singh, Fen Xie, Matei Zaharia, Richard Zang,
  Juntai Zheng, and Corey Zumar.
\newblock Developments in mlflow: {A} system to accelerate the machine learning
  lifecycle.
\newblock In {\em DEEM@SIGMOD}, pages 5:1--5:4, 2020.

\bibitem{DBLP:conf/nips/ChenJS18}
Irene~Y. Chen, Fredrik~D. Johansson, and David~A. Sontag.
\newblock Why is my classifier discriminatory?
\newblock In {\em NeurIPS}, pages 3543--3554, 2018.

\bibitem{ChenG16}
Tianqi Chen and Carlos Guestrin.
\newblock Xgboost: {A} scalable tree boosting system.
\newblock In {\em KDD}, pages 785--794, 2016.

\bibitem{cheng2019high}
Yu~Cheng, Ilias Diakonikolas, and Rong Ge.
\newblock High-dimensional robust mean estimation in nearly-linear time.
\newblock In {\em SIAM}, pages 2755--2771, 2019.

\bibitem{DBLP:conf/icml/ChoiGSSE20}
Kristy Choi, Aditya Grover, Trisha Singh, Rui Shu, and Stefano Ermon.
\newblock Fair generative modeling via weak supervision.
\newblock In {\em ICML}, pages 1887--1898, 2020.

\bibitem{DBLP:journals/bigdata/Chouldechova17}
Alexandra Chouldechova.
\newblock Fair prediction with disparate impact: {A} study of bias in
  recidivism prediction instruments.
\newblock {\em Big Data}, 5(2):153--163, 2017.

\bibitem{DBLP:journals/cacm/ChouldechovaR20}
Alexandra Chouldechova and Aaron Roth.
\newblock A snapshot of the frontiers of fairness in machine learning.
\newblock {\em Commun. {ACM}}, 63(5):82--89, 2020.

\bibitem{DBLP:conf/nips/ChzhenDHOP19}
Evgenii Chzhen, Christophe Denis, Mohamed Hebiri, Luca Oneto, and Massimiliano
  Pontil.
\newblock Leveraging labeled and unlabeled data for consistent fair binary
  classification.
\newblock In {\em NeurIPS}, pages 12739--12750, 2019.

\bibitem{DBLP:conf/alt/CotterJS19}
Andrew Cotter, Heinrich Jiang, and Karthik Sridharan.
\newblock Two-player games for efficient non-convex constrained optimization.
\newblock In {\em ALT}, pages 300--332, 2019.

\bibitem{DBLP:conf/sp/CretuSLSK08}
Gabriela~F. Cretu, Angelos Stavrou, Michael~E. Locasto, Salvatore~J. Stolfo,
  and Angelos~D. Keromytis.
\newblock Casting out demons: Sanitizing training data for anomaly sensors.
\newblock In {\em IEEE S{\&}P}, pages 81--95, 2008.

\bibitem{DBLP:conf/cvpr/CubukZMVL19}
Ekin~D. Cubuk, Barret Zoph, Dandelion Man{\'{e}}, Vijay Vasudevan, and Quoc~V.
  Le.
\newblock Autoaugment: Learning augmentation strategies from data.
\newblock In {\em CVPR}, pages 113--123, 2019.

\bibitem{diakonikolas2019robust}
Ilias Diakonikolas, Gautam Kamath, Daniel Kane, Jerry Li, Ankur Moitra, and
  Alistair Stewart.
\newblock Robust estimators in high-dimensions without the computational
  intractability.
\newblock {\em SIAM Journal on Computing}, 48(2):742--864, 2019.

\bibitem{2016COMPASRS}
William Dieterich, Christina Mendoza, and Tim Brennan.
\newblock Compas risk scales: Demonstrating accuracy equity and predictive
  parity.
\newblock Technical report, Northpoint Inc, 2016.

\bibitem{DBLP:books/daglib/0029346}
AnHai Doan, Alon~Y. Halevy, and Zachary~G. Ives.
\newblock {\em Principles of Data Integration}.
\newblock Morgan Kaufmann, 2012.

\bibitem{DBLP:journals/pvldb/DolatshahTWP18}
Mohamad Dolatshah, Mathew Teoh, Jiannan Wang, and Jian Pei.
\newblock Cleaning crowdsourced labels using oracles for statistical
  classification.
\newblock {\em PVLDB}, 12(4):376--389, 2018.

\bibitem{DBLP:conf/kdd/DongR19}
Xin~Luna Dong and Theodoros Rekatsinas.
\newblock Data integration and machine learning: {A} natural synergy.
\newblock In {\em KDD}, pages 3193--3194, 2019.

\bibitem{DBLP:journals/debu/DrevesHPPRC21}
Mike Dreves, Gene Huang, Zhuo Peng, Neoklis Polyzotis, Evan Rosen, and Paul
  Suganthan~G. C.
\newblock Validating data and models in continuous {ML} pipelines.
\newblock {\em {IEEE} Data Eng. Bull.}, 44(1):42--50, 2021.

\bibitem{Dua:2019}
Dheeru Dua and Casey Graff.
\newblock {UCI} machine learning repository, 2017.

\bibitem{DBLP:conf/innovations/DworkHPRZ12}
Cynthia Dwork, Moritz Hardt, Toniann Pitassi, Omer Reingold, and Richard~S.
  Zemel.
\newblock Fairness through awareness.
\newblock In {\em ITCS}, pages 214--226, 2012.

\bibitem{DBLP:conf/kdd/FeldmanFMSV15}
Michael Feldman, Sorelle~A. Friedler, John Moeller, Carlos Scheidegger, and
  Suresh Venkatasubramanian.
\newblock Certifying and removing disparate impact.
\newblock In {\em KDD}, pages 259--268, 2015.

\bibitem{DBLP:conf/icde/FernandezAKYMS18}
Raul~Castro Fernandez, Ziawasch Abedjan, Famien Koko, Gina Yuan, Samuel Madden,
  and Michael Stonebraker.
\newblock Aurum: {A} data discovery system.
\newblock In {\em ICDE}, pages 1001--1012, 2018.

\bibitem{https://doi.org/10.1111/j.1467-9868.2007.00643.x}
Dean~P. Foster and Robert~A. Stine.
\newblock Alpha-investing: a procedure for sequential control of expected false
  discoveries.
\newblock {\em Journal of the Royal Statistical Society: Series B (Statistical
  Methodology)}, 70(2):429--444, 2008.

\bibitem{modelpatching}
Karan Goel, Albert Gu, Yixuan Li, and Christopher R{\'{e}}.
\newblock Model patching: Closing the subgroup performance gap with data
  augmentation.
\newblock In {\em ICLR}, 2021.

\bibitem{DBLP:journals/corr/Goodfellow17}
Ian~J. Goodfellow.
\newblock {NIPS} 2016 tutorial: Generative adversarial networks.
\newblock {\em CoRR}, abs/1701.00160, 2017.

\bibitem{DBLP:conf/nips/GoodfellowPMXWOCB14}
Ian~J. Goodfellow, Jean Pouget{-}Abadie, Mehdi Mirza, Bing Xu, David
  Warde{-}Farley, Sherjil Ozair, Aaron~C. Courville, and Yoshua Bengio.
\newblock Generative adversarial nets.
\newblock In {\em NIPS}, pages 2672--2680, 2014.

\bibitem{DBLP:journals/corr/GoodfellowSS14}
Ian~J. Goodfellow, Jonathon Shlens, and Christian Szegedy.
\newblock Explaining and harnessing adversarial examples.
\newblock In {\em {ICLR}}, 2015.

\bibitem{tfhub}
J.~Gordon.
\newblock Introducing tensorflow hub: A library for reusable machine learning
  modules in tensorflow., 2018.

\bibitem{DBLP:conf/cidr/GrafbergerSS21}
Stefan Grafberger, Julia Stoyanovich, and Sebastian Schelter.
\newblock Lightweight inspection of data preprocessing in native machine
  learning pipelines.
\newblock In {\em CIDR}, 2021.

\bibitem{DBLP:conf/sigmod/HalevyKNOPRW16}
Alon~Y. Halevy, Flip Korn, Natalya~Fridman Noy, Christopher Olston, Neoklis
  Polyzotis, Sudip Roy, and Steven~Euijong Whang.
\newblock Goods: Organizing {G}oogle's datasets.
\newblock In {\em SIGMOD}, pages 795--806, 2016.

\bibitem{DBLP:conf/nips/HanYYNXHTS18}
Bo~Han, Quanming Yao, Xingrui Yu, Gang Niu, Miao Xu, Weihua Hu, Ivor~W. Tsang,
  and Masashi Sugiyama.
\newblock Co-teaching: Robust training of deep neural networks with extremely
  noisy labels.
\newblock In {\em NeurIPS}, pages 8536--8546, 2018.

\bibitem{DBLP:conf/nips/HardtPNS16}
Moritz Hardt, Eric Price, and Nati Srebro.
\newblock Equality of opportunity in supervised learning.
\newblock In {\em NIPS}, pages 3315--3323, 2016.

\bibitem{DBLP:conf/icml/HashimotoSNL18}
Tatsunori~B. Hashimoto, Megha Srivastava, Hongseok Namkoong, and Percy Liang.
\newblock Fairness without demographics in repeated loss minimization.
\newblock In Jennifer~G. Dy and Andreas Krause, editors, {\em ICML}, volume~80,
  pages 1934--1943. {PMLR}, 2018.

\bibitem{DBLP:conf/hpca/HazelwoodBBCDDF18}
Kim~M. Hazelwood, Sarah Bird, David~M. Brooks, Soumith Chintala, Utku Diril,
  Dmytro Dzhulgakov, Mohamed Fawzy, Bill Jia, Yangqing Jia, Aditya Kalro, James
  Law, Kevin Lee, Jason Lu, Pieter Noordhuis, Misha Smelyanskiy, Liang Xiong,
  and Xiaodong Wang.
\newblock Applied machine learning at facebook: {A} datacenter infrastructure
  perspective.
\newblock In {\em HPCA}, pages 620--629, 2018.

\bibitem{DBLP:conf/iclr/HendrycksMCZGL20}
Dan Hendrycks, Norman Mu, Ekin~Dogus Cubuk, Barret Zoph, Justin Gilmer, and
  Balaji Lakshminarayanan.
\newblock Augmix: {A} simple data processing method to improve robustness and
  uncertainty.
\newblock In {\em ICLR}, 2020.

\bibitem{DBLP:journals/corr/abs-2004-03264}
Geon Heo, Yuji Roh, Seonghyeon Hwang, Dayun Lee, and Steven~Euijong Whang.
\newblock Inspector gadget: {A} data programming-based labeling system for
  industrial images.
\newblock {\em {PVLDB}}, 2021.

\bibitem{michelangelo}
M.~Hermann, J.and Del~Baso.
\newblock Meet michelangelo: Uber’s machine learning platform, 2017.

\bibitem{DBLP:journals/air/HodgeA04}
Victoria~J. Hodge and Jim Austin.
\newblock A survey of outlier detection methodologies.
\newblock {\em Artif. Intell. Rev.}, 22(2):85--126, 2004.

\bibitem{huber1992robust}
Peter~J Huber.
\newblock Robust estimation of a location parameter.
\newblock In {\em Breakthroughs in statistics}, pages 492--518. Springer, 1992.

\bibitem{DBLP:books/acm/IlyasC19}
Ihab~F. Ilyas and Xu~Chu.
\newblock {\em Data Cleaning}.
\newblock {ACM}, 2019.

\bibitem{10.1145/3506712}
Ihab~F. Ilyas and Theodoros Rekatsinas.
\newblock Machine learning and data cleaning: Which serves the other?
\newblock {\em J. Data and Information Quality}, dec 2021.
\newblock Just Accepted.

\bibitem{DBLP:conf/cikm/IosifidisN19}
Vasileios Iosifidis and Eirini Ntoutsi.
\newblock Adafair: Cumulative fairness adaptive boosting.
\newblock In {\em CIKM}, pages 781--790, 2019.

\bibitem{DBLP:conf/aistats/JiangN20}
Heinrich Jiang and Ofir Nachum.
\newblock Identifying and correcting label bias in machine learning.
\newblock In {\em AISTATS}, pages 702--712, 2020.

\bibitem{DBLP:conf/icml/JiangZLLF18}
Lu~Jiang, Zhengyuan Zhou, Thomas Leung, Li{-}Jia Li, and Li~Fei{-}Fei.
\newblock Mentornet: Learning data-driven curriculum for very deep neural
  networks on corrupted labels.
\newblock In {\em ICML}, pages 2309--2318, 2018.

\bibitem{DBLP:journals/kais/KamiranC11}
Faisal Kamiran and Toon Calders.
\newblock Data preprocessing techniques for classification without
  discrimination.
\newblock {\em Knowl. Inf. Syst.}, 33(1):1--33, 2011.

\bibitem{DBLP:conf/pkdd/KamishimaAAS12}
Toshihiro Kamishima, Shotaro Akaho, Hideki Asoh, and Jun Sakuma.
\newblock Fairness-aware classifier with prejudice remover regularizer.
\newblock In {\em {ECML} {PKDD}}, pages 35--50, 2012.

\bibitem{DBLP:journals/pvldb/KarlasLWGC0020}
Bojan Karlas, Peng Li, Renzhi Wu, Nezihe~Merve G{\"{u}}rel, Xu~Chu, Wentao Wu,
  and Ce~Zhang.
\newblock Nearest neighbor classifiers over incomplete information: From
  certain answers to certain predictions.
\newblock {\em Proc. {VLDB} Endow.}, 14(3):255--267, 2020.

\bibitem{10.1145/3308558.3313559}
Aria Khademi, Sanghack Lee, David Foley, and Vasant Honavar.
\newblock Fairness in algorithmic decision making: An excursion through the
  lens of causality.
\newblock In {\em WWW}, pages 2907--2914, 2019.

\bibitem{DBLP:conf/fat/KhaniL21}
Fereshte Khani and Percy Liang.
\newblock Removing spurious features can hurt accuracy and affect groups
  disproportionately.
\newblock In {\em FAccT}, pages 196--205. {ACM}, 2021.

\bibitem{10.5555/3294771.3294834}
Niki Kilbertus, Mateo Rojas-Carulla, Giambattista Parascandolo, Moritz Hardt,
  Dominik Janzing, and Bernhard Sch\"{o}lkopf.
\newblock Avoiding discrimination through causal reasoning.
\newblock In {\em NeurIPS}, pages 656--666, 2017.

\bibitem{DBLP:conf/aaai/KimLHS19}
Hoon Kim, Kangwook Lee, Gyeongjo Hwang, and Changho Suh.
\newblock Crash to not crash: Learn to identify dangerous vehicles using a
  simulator.
\newblock In {\em AAAI}, pages 978--985, 2019.

\bibitem{DBLP:journals/corr/abs-1811-00741}
Pang~Wei Koh, Jacob Steinhardt, and Percy Liang.
\newblock Stronger data poisoning attacks break data sanitization defenses.
\newblock {\em CoRR}, abs/1811.00741, 2018.

\bibitem{DBLP:journals/pvldb/KrishnanWWFG16}
Sanjay Krishnan, Jiannan Wang, Eugene Wu, Michael~J. Franklin, and Ken
  Goldberg.
\newblock Activeclean: Interactive data cleaning for statistical modeling.
\newblock {\em PVLDB}, 9(12):948--959, 2016.

\bibitem{DBLP:journals/corr/abs-1807-04720}
Karol Kurach, Mario Lucic, Xiaohua Zhai, Marcin Michalski, and Sylvain Gelly.
\newblock The {GAN} landscape: Losses, architectures, regularization, and
  normalization.
\newblock {\em CoRR}, abs/1807.04720, 2018.

\bibitem{NIPS2017_6995}
Matt~J Kusner, Joshua Loftus, Chris Russell, and Ricardo Silva.
\newblock Counterfactual fairness.
\newblock In {\em NeurIPS}, pages 4066--4076. 2017.

\bibitem{DBLP:conf/nips/LahotiBCLPT0C20}
Preethi Lahoti, Alex Beutel, Jilin Chen, Kang Lee, Flavien Prost, Nithum Thain,
  Xuezhi Wang, and Ed~Chi.
\newblock Fairness without demographics through adversarially reweighted
  learning.
\newblock In {\em NeurIPS}, 2020.

\bibitem{DBLP:conf/nips/LamyZ19}
Alexandre~Louis Lamy and Ziyuan Zhong.
\newblock Noise-tolerant fair classification.
\newblock In Hanna~M. Wallach, Hugo Larochelle, Alina Beygelzimer, Florence
  d'Alch{\'{e}}{-}Buc, Emily~B. Fox, and Roman Garnett, editors, {\em NeurIPS},
  pages 294--305, 2019.

\bibitem{DBLP:journals/debu/LeeP18}
Doris Jung~Lin Lee and Aditya~G. Parameswaran.
\newblock The case for a visual discovery assistant: {A} holistic solution for
  accelerating visual data exploration.
\newblock {\em {IEEE} Data Eng. Bull.}, 41(3):3--14, 2018.

\bibitem{DBLP:conf/kdd/0001RSW21}
Jae{-}Gil Lee, Yuji Roh, Hwanjun Song, and Steven~Euijong Whang.
\newblock Machine learning robustness, fairness, and their convergence.
\newblock In {\em KDD}, pages 4046--4047, 2021.

\bibitem{DBLP:conf/iclr/LiSH20}
Junnan Li, Richard Socher, and Steven C.~H. Hoi.
\newblock Dividemix: Learning with noisy labels as semi-supervised learning.
\newblock In {\em ICLR}, 2020.

\bibitem{cleanmlli}
Peng Li, Xi~Rao, Jennifer Blase, Yue Zhang, Xu~Chu, and Ce~Zhang.
\newblock {CleanML}: {A} benchmark for joint data cleaning and machine learning
  [experiments and analysis].
\newblock In {\em ICDE}, 2021.

\bibitem{liu2021robust}
Zifan Liu, Jong~Ho Park, Theodoros Rekatsinas, and Christos Tzamos.
\newblock On robust mean estimation under coordinate-level corruption.
\newblock In {\em ICML}, pages 6914--6924. PMLR, 2021.

\bibitem{DBLP:conf/icml/LiuPRT21}
Zifan Liu, Jongho Park, Theodoros Rekatsinas, and Christos Tzamos.
\newblock On robust mean estimation under coordinate-level corruption.
\newblock In {\em ICML}, pages 6914--6924, 2021.

\bibitem{DBLP:conf/nips/MalachS17}
Eran Malach and Shai Shalev{-}Shwartz.
\newblock Decoupling "when to update" from "how to update".
\newblock In {\em NIPS}, pages 960--970, 2017.

\bibitem{DBLP:journals/corr/abs-1908-09635}
Ninareh Mehrabi, Fred Morstatter, Nripsuta Saxena, Kristina Lerman, and Aram
  Galstyan.
\newblock A survey on bias and fairness in machine learning.
\newblock {\em CoRR}, abs/1908.09635, 2019.

\bibitem{DBLP:conf/cidr/MelgarDGGHJKL0L21}
Leonel~Aguilar Melgar, David Dao, Shaoduo Gan, Nezihe~Merve G{\"{u}}rel, Nora
  Hollenstein, Jiawei Jiang, Bojan Karlas, Thomas Lemmin, Tian Li, Yang Li,
  Xi~Rao, Johannes Rausch, C{\'{e}}dric Renggli, Luka Rimanic, Maurice Weber,
  Shuai Zhang, Zhikuan Zhao, Kevin Schawinski, Wentao Wu, and Ce~Zhang.
\newblock Ease.ml: {A} lifecycle management system for machine learning.
\newblock In {\em CIDR}, 2021.

\bibitem{DBLP:conf/ccs/MengC17}
Dongyu Meng and Hao Chen.
\newblock Magnet: {A} two-pronged defense against adversarial examples.
\newblock In Bhavani~M. Thuraisingham, David Evans, Tal Malkin, and Dongyan Xu,
  editors, {\em {ACM} {SIGSAC}}, pages 135--147, 2017.

\bibitem{DBLP:conf/iclr/MetzenGFB17}
Jan~Hendrik Metzen, Tim Genewein, Volker Fischer, and Bastian Bischoff.
\newblock On detecting adversarial perturbations.
\newblock In {\em {ICLR}}, 2017.

\bibitem{DBLP:journals/debu/MillerNZCPA18}
Ren{\'{e}}e~J. Miller, Fatemeh Nargesian, Erkang Zhu, Christina
  Christodoulakis, Ken~Q. Pu, and Periklis Andritsos.
\newblock Making open data transparent: Data discovery on open data.
\newblock {\em {IEEE} Data Eng. Bull.}, 41(2):59--70, 2018.

\bibitem{DBLP:conf/acl/MintzBSJ09}
Mike Mintz, Steven Bills, Rion Snow, and Daniel Jurafsky.
\newblock Distant supervision for relation extraction without labeled data.
\newblock In Keh{-}Yih Su, Jian Su, and Janyce Wiebe, editors, {\em ACL}, pages
  1003--1011, 2009.

\bibitem{Nabi2018FairIO}
Razieh Nabi and Ilya Shpitser.
\newblock Fair inference on outcomes.
\newblock In {\em AAAI}, pages 1931--1940, 2018.

\bibitem{DBLP:journals/debu/NeutatzCA021}
Felix Neutatz, Binger Chen, Ziawasch Abedjan, and Eugene Wu.
\newblock From cleaning before {ML} to cleaning for {ML}.
\newblock {\em {IEEE} Data Eng. Bull.}, 44(1):24--41, 2021.

\bibitem{DBLP:conf/eccv/NorooziF16}
Mehdi Noroozi and Paolo Favaro.
\newblock Unsupervised learning of visual representations by solving jigsaw
  puzzles.
\newblock In {\em ECCV}, pages 69--84, 2016.

\bibitem{DBLP:journals/tkde/PanY10}
Sinno~Jialin Pan and Qiang Yang.
\newblock A survey on transfer learning.
\newblock {\em {IEEE} TKDE}, 22(10):1345--1359, 2010.

\bibitem{DBLP:conf/sp/PapernotM0JS16}
Nicolas Papernot, Patrick~D. McDaniel, Xi~Wu, Somesh Jha, and Ananthram Swami.
\newblock Distillation as a defense to adversarial perturbations against deep
  neural networks.
\newblock In {\em {IEEE} {SP}}, pages 582--597, 2016.

\bibitem{NEURIPS2019_9015}
Adam Paszke, Sam Gross, Francisco Massa, Adam Lerer, James Bradbury, Gregory
  Chanan, Trevor Killeen, Zeming Lin, Natalia Gimelshein, Luca Antiga, Alban
  Desmaison, Andreas Kopf, Edward Yang, Zachary DeVito, Martin Raison, Alykhan
  Tejani, Sasank Chilamkurthy, Benoit Steiner, Lu~Fang, Junjie Bai, and Soumith
  Chintala.
\newblock Pytorch: An imperative style, high-performance deep learning library.
\newblock In H.~Wallach, H.~Larochelle, A.~Beygelzimer, F.~d\textquotesingle
  Alch\'{e}-Buc, E.~Fox, and R.~Garnett, editors, {\em NeurIPS}, pages
  8024--8035. Curran Associates, Inc., 2019.

\bibitem{DBLP:conf/cvpr/PatriniRMNQ17}
Giorgio Patrini, Alessandro Rozza, Aditya~Krishna Menon, Richard Nock, and
  Lizhen Qu.
\newblock Making deep neural networks robust to label noise: {A} loss
  correction approach.
\newblock In {\em CVPR}, pages 2233--2241, 2017.

\bibitem{DBLP:journals/corr/abs-1802-03041}
Andrea Paudice, Luis Mu{\~{n}}oz{-}Gonz{\'{a}}lez, Andr{\'{a}}s Gy{\"{o}}rgy,
  and Emil~C. Lupu.
\newblock Detection of adversarial training examples in poisoning attacks
  through anomaly detection.
\newblock {\em CoRR}, abs/1802.03041, 2018.

\bibitem{DBLP:conf/pkdd/PelekisNTST13}
Nikos Pelekis, Christos Ntrigkogias, Panagiotis Tampakis, Stylianos Sideridis,
  and Yannis Theodoridis.
\newblock Hermoupolis: {A} trajectory generator for simulating generalized
  mobility patterns.
\newblock In {\em {ECML} {PKDD}}, pages 659--662, 2013.

\bibitem{DBLP:conf/nips/PleissRWKW17}
Geoff Pleiss, Manish Raghavan, Felix Wu, Jon~M. Kleinberg, and Kilian~Q.
  Weinberger.
\newblock On fairness and calibration.
\newblock In {\em NIPS}, pages 5680--5689, 2017.

\bibitem{Polyzotis:2017:DMC:3035918.3054782}
Neoklis Polyzotis, Sudip Roy, Steven~Euijong Whang, and Martin Zinkevich.
\newblock Data management challenges in production machine learning.
\newblock In {\em SIGMOD}, pages 1723--1726, 2017.

\bibitem{DBLP:journals/sigmod/PolyzotisRWZ18}
Neoklis Polyzotis, Sudip Roy, Steven~Euijong Whang, and Martin Zinkevich.
\newblock Data lifecycle challenges in production machine learning: {A} survey.
\newblock {\em {SIGMOD} Rec.}, 47(2):17--28, 2018.

\bibitem{qayyum2020secure}
Adnan Qayyum, Junaid Qadir, Muhammad Bilal, and Ala Al-Fuqaha.
\newblock Secure and robust machine learning for healthcare: A survey.
\newblock {\em IEEE Reviews in Biomedical Engineering}, 14:156--180, 2020.

\bibitem{Ratner:2017:SRT:3173074.3173077}
Alexander Ratner, Stephen~H. Bach, Henry Ehrenberg, Jason Fries, Sen Wu, and
  Christopher R{\'e}.
\newblock Snorkel: Rapid training data creation with weak supervision.
\newblock {\em PVLDB}, 11(3):269--282, November 2017.

\bibitem{DBLP:journals/vldb/RatnerBEFWR20}
Alexander Ratner, Stephen~H. Bach, Henry~R. Ehrenberg, Jason~A. Fries, Sen Wu,
  and Christopher R{\'{e}}.
\newblock Snorkel: rapid training data creation with weak supervision.
\newblock {\em {VLDB} J.}, 29(2-3):709--730, 2020.

\bibitem{DBLP:conf/nips/RatnerEHDR17}
Alexander~J. Ratner, Henry~R. Ehrenberg, Zeshan Hussain, Jared Dunnmon, and
  Christopher R{\'{e}}.
\newblock Learning to compose domain-specific transformations for data
  augmentation.
\newblock In {\em NIPS}, pages 3239--3249, 2017.

\bibitem{DBLP:conf/edbt/RedyukKMS21}
Sergey Redyuk, Zoi Kaoudi, Volker Markl, and Sebastian Schelter.
\newblock Automating data quality validation for dynamic data ingestion.
\newblock In {\em EDBT}, pages 61--72, 2021.

\bibitem{DBLP:journals/corr/ReedLASER14}
Scott~E. Reed, Honglak Lee, Dragomir Anguelov, Christian Szegedy, Dumitru
  Erhan, and Andrew Rabinovich.
\newblock Training deep neural networks on noisy labels with bootstrapping.
\newblock In {\em ICLR}, 2015.

\bibitem{DBLP:journals/pvldb/RekatsinasCIR17}
Theodoros Rekatsinas, Xu~Chu, Ihab~F. Ilyas, and Christopher R{\'{e}}.
\newblock Holoclean: Holistic data repairs with probabilistic inference.
\newblock {\em PVLDB}, 10(11):1190--1201, 2017.

\bibitem{DBLP:journals/debu/RenggliRGK0021}
C{\'{e}}dric Renggli, Luka Rimanic, Nezihe~Merve G{\"{u}}rel, Bojan Karlas,
  Wentao Wu, and Ce~Zhang.
\newblock A data quality-driven view of mlops.
\newblock {\em {IEEE} Data Eng. Bull.}, 44(1):11--23, 2021.

\bibitem{DBLP:reference/sp/2015rsh}
Francesco Ricci, Lior Rokach, and Bracha Shapira, editors.
\newblock {\em Recommender Systems Handbook}.
\newblock Springer, 2015.

\bibitem{DBLP:journals/tkde/RohHW19}
Yuji Roh, Geon Heo, and Steven~Euijong Whang.
\newblock A survey on data collection for machine learning: a big data - {AI}
  integration perspective.
\newblock {\em {IEEE} TKDE}, 2019.

\bibitem{DBLP:journals/corr/abs-2002-10234}
Yuji Roh, Kangwook Lee, Steven~Euijong Whang, and Changho Suh.
\newblock {FR-Train}: {A} mutual information-based approach to fair and robust
  training.
\newblock In {\em ICML}, 2020.

\bibitem{DBLP:conf/iclr/Roh0WS21}
Yuji Roh, Kangwook Lee, Steven~Euijong Whang, and Changho Suh.
\newblock Fairbatch: Batch selection for model fairness.
\newblock In {\em ICLR}. OpenReview.net, 2021.

\bibitem{roh2021sample}
Yuji Roh, Kangwook Lee, Steven~Euijong Whang, and Changho Suh.
\newblock Sample selection for fair and robust training.
\newblock In {\em NeurIPS}, 2021.

\bibitem{DBLP:conf/sigmod/SalimiRHS19}
Babak Salimi, Luke Rodriguez, Bill Howe, and Dan Suciu.
\newblock Interventional fairness: Causal database repair for algorithmic
  fairness.
\newblock In {\em SIGMOD}, pages 793--810, 2019.

\bibitem{Schelter2017AutomaticallyTM}
S.~Schelter, Joos-Hendrik B{\"o}se, Johannes Kirschnick, T.~Klein, and Stephan
  Seufert.
\newblock Automatically tracking metadata and provenance of machine learning
  experiments.
\newblock In {\em Workshop on ML Systems at NIPS}, 2017.

\bibitem{DBLP:conf/icde/SchelterGSRKTBL19}
Sebastian Schelter, Stefan Grafberger, Philipp Schmidt, Tammo Rukat, Mario
  Kie{\ss}ling, Andrey Taptunov, Felix Bie{\ss}mann, and Dustin Lange.
\newblock Differential data quality verification on partitioned data.
\newblock In {\em ICDE}, pages 1940--1945, 2019.

\bibitem{DBLP:journals/pvldb/SchelterLSCBG18}
Sebastian Schelter, Dustin Lange, Philipp Schmidt, Meltem Celikel, Felix
  Bie{\ss}mann, and Andreas Grafberger.
\newblock Automating large-scale data quality verification.
\newblock {\em Proc. {VLDB} Endow.}, 11(12):1781--1794, 2018.

\bibitem{DBLP:conf/edbt/SchelterRB21}
Sebastian Schelter, Tammo Rukat, and Felix Biessmann.
\newblock {JENGA} - {A} framework to study the impact of data errors on the
  predictions of machine learning models.
\newblock In {\em EDBT}, pages 529--534, 2021.

\bibitem{DBLP:conf/nips/SculleyHGDPECYC15}
D.~Sculley, Gary Holt, Daniel Golovin, Eugene Davydov, Todd Phillips, Dietmar
  Ebner, Vinay Chaudhary, Michael Young, Jean{-}Fran{\c{c}}ois Crespo, and Dan
  Dennison.
\newblock Hidden technical debt in machine learning systems.
\newblock In {\em NIPS}, pages 2503--2511, 2015.

\bibitem{DBLP:series/synthesis/2012Settles}
Burr Settles.
\newblock {\em Active Learning}.
\newblock Synthesis Lectures on Artificial Intelligence and Machine Learning.
  Morgan {\&} Claypool Publishers, 2012.

\bibitem{DBLP:conf/nips/ShafahiHNSSDG18}
Ali Shafahi, W.~Ronny Huang, Mahyar Najibi, Octavian Suciu, Christoph Studer,
  Tudor Dumitras, and Tom Goldstein.
\newblock Poison frogs! targeted clean-label poisoning attacks on neural
  networks.
\newblock In {\em NeurIPS}, pages 6106--6116, 2018.

\bibitem{shang2008denoising}
Li~Shang.
\newblock Denoising natural images based on a modified sparse coding algorithm.
\newblock {\em Applied mathematics and computation}, 205(2):883--889, 2008.

\bibitem{shen2021towards}
Zheyan Shen, Jiashuo Liu, Yue He, Xingxuan Zhang, Renzhe Xu, Han Yu, and Peng
  Cui.
\newblock Towards out-of-distribution generalization: A survey.
\newblock {\em arXiv preprint arXiv:2108.13624}, 2021.

\bibitem{DBLP:conf/kdd/ShengPI08}
Victor~S. Sheng, Foster~J. Provost, and Panagiotis~G. Ipeirotis.
\newblock Get another label? improving data quality and data mining using
  multiple, noisy labelers.
\newblock In {\em KDD}, pages 614--622, 2008.

\bibitem{Sinha2017CertifyingSD}
Aman Sinha, Hongseok Namkoong, and John~C. Duchi.
\newblock Certifying some distributional robustness with principled adversarial
  training.
\newblock In {\em ICLR}, 2018.

\bibitem{DBLP:conf/pkdd/SolansB020}
David Solans, Battista Biggio, and Carlos Castillo.
\newblock Poisoning attacks on algorithmic fairness.
\newblock In Frank Hutter, Kristian Kersting, Jefrey Lijffijt, and Isabel
  Valera, editors, {\em {ECML} {PKDD}}, volume 12457, pages 162--177. Springer,
  2020.

\bibitem{DBLP:conf/icml/SongK019}
Hwanjun Song, Minseok Kim, and Jae{-}Gil Lee.
\newblock {SELFIE:} refurbishing unclean samples for robust deep learning.
\newblock In {\em ICML}, pages 5907--5915, 2019.

\bibitem{DBLP:journals/corr/abs-2007-08199}
Hwanjun Song, Minseok Kim, Dongmin Park, and Jae{-}Gil Lee.
\newblock Learning from noisy labels with deep neural networks: {A} survey.
\newblock {\em CoRR}, abs/2007.08199, 2020.

\bibitem{DBLP:conf/kdd/SongKPS021}
Hwanjun Song, Minseok Kim, Dongmin Park, Yooju Shin, and Jae{-}Gil Lee.
\newblock Robust learning by self-transition for handling noisy labels.
\newblock In {\em {KDD}}, pages 1490--1500, 2021.

\bibitem{DBLP:journals/debu/StonebrakerI18}
Michael Stonebraker and Ihab~F. Ilyas.
\newblock Data integration: The current status and the way forward.
\newblock {\em {IEEE} Data Eng. Bull.}, 41(2):3--9, 2018.

\bibitem{DBLP:journals/debu/Stonebraker19}
Michael Stonebraker and El~Kindi Rezig.
\newblock Machine learning and big data: What is important?
\newblock {\em {IEEE} Data Eng. Bull.}, 2019.

\bibitem{DBLP:conf/sigmod/TaeW21}
Ki~Hyun Tae and Steven~Euijong Whang.
\newblock Slice tuner: {A} selective data acquisition framework for accurate
  and fair machine learning models.
\newblock In {\em SIGMOD}, pages 1771--1783. {ACM}, 2021.

\bibitem{DBLP:conf/nips/TarvainenV17}
Antti Tarvainen and Harri Valpola.
\newblock Mean teachers are better role models: Weight-averaged consistency
  targets improve semi-supervised deep learning results.
\newblock In {\em NIPS}, pages 1195--1204, 2017.

\bibitem{DBLP:conf/cidr/TerrizzanoSRC15}
Ignacio~G. Terrizzano, Peter~M. Schwarz, Mary Roth, and John~E. Colino.
\newblock Data wrangling: The challenging yourney from the wild to the lake.
\newblock In {\em CIDR}, 2015.

\bibitem{DBLP:conf/iros/TobinFRSZA17}
Josh Tobin, Rachel Fong, Alex Ray, Jonas Schneider, Wojciech Zaremba, and
  Pieter Abbeel.
\newblock Domain randomization for transferring deep neural networks from
  simulation to the real world.
\newblock In {\em IROS}, pages 23--30, 2017.

\bibitem{DBLP:conf/cvpr/TremblayPABJATC18}
Jonathan Tremblay, Aayush Prakash, David Acuna, Mark Brophy, Varun Jampani, Cem
  Anil, Thang To, Eric Cameracci, Shaad Boochoon, and Stan Birchfield.
\newblock Training deep networks with synthetic data: Bridging the reality gap
  by domain randomization.
\newblock In {\em {CVPR} Workshops}, pages 969--977, 2018.

\bibitem{DBLP:journals/kais/TrigueroGH15}
Isaac Triguero, Salvador Garc{\'{\i}}a, and Francisco Herrera.
\newblock Self-labeled techniques for semi-supervised learning: taxonomy,
  software and empirical study.
\newblock {\em Knowl. Inf. Syst.}, 42(2):245--284, 2015.

\bibitem{tukey1960survey}
John~W Tukey.
\newblock A survey of sampling from contaminated distributions.
\newblock {\em Contributions to probability and statistics}, pages 448--485,
  1960.

\bibitem{JSSv045i03}
Stef van Buuren and Karin Groothuis-Oudshoorn.
\newblock mice: Multivariate imputation by chained equations in r.
\newblock {\em Journal of Statistical Software}, 45(3):1–--67, 2011.

\bibitem{DBLP:journals/pvldb/VarmaR18}
Paroma Varma and Christopher R{\'{e}}.
\newblock Snuba: Automating weak supervision to label training data.
\newblock {\em Proc. {VLDB} Endow.}, 12(3):223--236, 2018.

\bibitem{DBLP:journals/pvldb/VartakRMPP15}
Manasi Vartak, Sajjadur Rahman, Samuel Madden, Aditya~G. Parameswaran, and
  Neoklis Polyzotis.
\newblock {SEEDB}: Efficient data-driven visualization recommendations to
  support visual analytics.
\newblock {\em {PVLDB}}, 8(13):2182--2193, 2015.

\bibitem{DBLP:conf/pods/Venkatasubramanian19}
Suresh Venkatasubramanian.
\newblock Algorithmic fairness: Measures, methods and representations.
\newblock In {\em PODS}, page 481, 2019.

\bibitem{wang2019learning}
Hao Wang, Bing Liu, Chaozhuo Li, Yan Yang, and Tianrui Li.
\newblock Learning with noisy labels for sentence-level sentiment
  classification.
\newblock In {\em EMNLP}, 2019.

\bibitem{DBLP:conf/fat/WangLL21}
Jialu Wang, Yang Liu, and Caleb Levy.
\newblock Fair classification with group-dependent label noise.
\newblock In Madeleine~Clare Elish, William Isaac, and Richard~S. Zemel,
  editors, {\em FAccT}, pages 526--536. {ACM}, 2021.

\bibitem{DBLP:conf/nips/WangGNCGJ20}
Serena Wang, Wenshuo Guo, Harikrishna Narasimhan, Andrew Cotter, Maya~R. Gupta,
  and Michael~I. Jordan.
\newblock Robust optimization for fairness with noisy protected groups.
\newblock In Hugo Larochelle, Marc'Aurelio Ranzato, Raia Hadsell,
  Maria{-}Florina Balcan, and Hsuan{-}Tien Lin, editors, {\em NeurIPS}, 2020.

\bibitem{DBLP:journals/pvldb/Whang020}
Steven~Euijong Whang and Jae{-}Gil Lee.
\newblock Data collection and quality challenges for deep learning.
\newblock {\em Proc. {VLDB} Endow.}, 13(12):3429--3432, 2020.

\bibitem{DBLP:journals/debu/XinPTWGHJP21}
Doris Xin, Devin Petersohn, Dixin Tang, Yifan Wu, Joseph~E. Gonzalez, Joseph~M.
  Hellerstein, Anthony~D. Joseph, and Aditya~G. Parameswaran.
\newblock Enhancing the interactivity of dataframe queries by leveraging think
  time.
\newblock {\em {IEEE} Data Eng. Bull.}, 44(1):66--78, 2021.

\bibitem{DBLP:conf/bigdataconf/XuYZW18}
Depeng Xu, Shuhan Yuan, Lu~Zhang, and Xintao Wu.
\newblock Fairgan: Fairness-aware generative adversarial networks.
\newblock In {\em {IEEE} Big Data}, pages 570--575, 2018.

\bibitem{DBLP:conf/icml/XuLLJT21}
Han Xu, Xiaorui Liu, Yaxin Li, Anil~K. Jain, and Jiliang Tang.
\newblock To be robust or to be fair: Towards fairness in adversarial training.
\newblock In Marina Meila and Tong Zhang, editors, {\em ICML}, volume 139,
  pages 11492--11501. {PMLR}, 2021.

\bibitem{DBLP:conf/ndss/Xu0Q18}
Weilin Xu, David Evans, and Yanjun Qi.
\newblock Feature squeezing: Detecting adversarial examples in deep neural
  networks.
\newblock In {\em {NDSS}}, 2018.

\bibitem{Yarowsky:1995:UWS:981658.981684}
David Yarowsky.
\newblock Unsupervised word sense disambiguation rivaling supervised methods.
\newblock In {\em ACL}, pages 189--196, Stroudsburg, PA, USA, 1995. Association
  for Computational Linguistics.

\bibitem{DBLP:conf/iccv/YunHCOYC19}
Sangdoo Yun, Dongyoon Han, Sanghyuk Chun, Seong~Joon Oh, Youngjoon Yoo, and
  Junsuk Choe.
\newblock Cutmix: Regularization strategy to train strong classifiers with
  localizable features.
\newblock In {\em ICCV}, pages 6022--6031, 2019.

\bibitem{DBLP:conf/www/ZafarVGG17}
Muhammad~Bilal Zafar, Isabel Valera, Manuel Gomez{-}Rodriguez, and Krishna~P.
  Gummadi.
\newblock Fairness beyond disparate treatment {\&} disparate impact: Learning
  classification without disparate mistreatment.
\newblock In {\em WWW}, pages 1171--1180. {ACM}, 2017.

\bibitem{DBLP:conf/aistats/ZafarVGG17}
Muhammad~Bilal Zafar, Isabel Valera, Manuel Gomez{-}Rodriguez, and Krishna~P.
  Gummadi.
\newblock Fairness constraints: Mechanisms for fair classification.
\newblock In {\em AISTATS}, pages 962--970, 2017.

\bibitem{DBLP:conf/aies/ZhangLM18}
Brian~Hu Zhang, Blake Lemoine, and Margaret Mitchell.
\newblock Mitigating unwanted biases with adversarial learning.
\newblock In {\em AIES}, pages 335--340, 2018.

\bibitem{DBLP:conf/sigmod/ZhangCAN21}
Hantian Zhang, Xu~Chu, Abolfazl Asudeh, and Shamkant~B. Navathe.
\newblock Omnifair: {A} declarative system for model-agnostic group fairness in
  machine learning.
\newblock In {\em SIGMOD}, pages 2076--2088, 2021.

\bibitem{DBLP:conf/fat/ZhangD21}
Hongjing Zhang and Ian Davidson.
\newblock Facct.
\newblock pages 138--148. {ACM}, 2021.

\bibitem{DBLP:conf/iclr/ZhangCDL18}
Hongyi Zhang, Moustapha Ciss{\'{e}}, Yann~N. Dauphin, and David Lopez{-}Paz.
\newblock mixup: Beyond empirical risk minimization.
\newblock In {\em ICLR}, 2018.

\bibitem{Zhang2018FairnessID}
Junzhe Zhang and Elias Bareinboim.
\newblock Fairness in decision-making - the causal explanation formula.
\newblock In {\em AAAI}, 2018.

\bibitem{DBLP:conf/sigmod/ZhangI20}
Yi~Zhang and Zachary~G. Ives.
\newblock Finding related tables in data lakes for interactive data science.
\newblock In {\em SIGMOD}, pages 1951--1966, 2020.

\bibitem{DBLP:conf/sigmod/ZhaoSZBUK17}
Zheguang Zhao, Lorenzo~De Stefani, Emanuel Zgraggen, Carsten Binnig, Eli Upfal,
  and Tim Kraska.
\newblock Controlling false discoveries during interactive data exploration.
\newblock In {\em SIGMOD}, pages 527--540, 2017.

\bibitem{DBLP:conf/ictai/ZhouG04}
Yan Zhou and Sally~A. Goldman.
\newblock Democratic co-learning.
\newblock In {\em IEEE ICTAI}, pages 594--602, 2004.

\bibitem{Zhou:2005:TEU:1092713.1092809}
Zhi-Hua Zhou and Ming Li.
\newblock Tri-training: Exploiting unlabeled data using three classifiers.
\newblock {\em IEEE TKDE}, 17(11):1529--1541, November 2005.

\bibitem{DBLP:conf/icml/ZhuHLTSG19}
Chen Zhu, W.~Ronny Huang, Hengduo Li, Gavin Taylor, Christoph Studer, and Tom
  Goldstein.
\newblock Transferable clean-label poisoning attacks on deep neural nets.
\newblock In {\em ICML}, pages 7614--7623, 2019.

\bibitem{zhu2008}
Xiaojin Zhu.
\newblock {Semi-Supervised Learning Literature Survey}.
\newblock Technical report, Computer Sciences, University of Wisconsin-Madison,
  2005.

\end{thebibliography}

\end{document}